\documentclass[10pt,journal,compsoc]{IEEEtran}

\ifCLASSOPTIONcompsoc
  \usepackage[nocompress]{cite}
\else
  \usepackage{cite}
\fi
\ifCLASSINFOpdf
\else
\fi

\usepackage[hang,flushmargin]{footmisc}

\usepackage{xurl}
\usepackage{colortbl}   
\usepackage{xcolor}  

\usepackage{tikz-dependency}
\usepackage{tikz}
\usepackage[edges]{forest}
\usepackage[numbers]{natbib}

\newcommand{\eg}{\textit{e.g.,}}
\definecolor{hidden-red}{RGB}{205, 44, 36}
\definecolor{hidden-blue}{RGB}{194,232,247}
\definecolor{hidden-orange}{RGB}{243,202,120}
\definecolor{hidden-green}{RGB}{34,139,34}
\definecolor{hidden-pink}{RGB}{255,245,247}
\definecolor{hidden-black}{RGB}{20,68,106}
\definecolor{purple}{RGB}{144,153,196}
\definecolor{yellow}{RGB}{255,228,123}
\definecolor{hidden-yellow}{RGB}{255,248,203}
\definecolor{tkcolor}{RGB}{224,223,255}
\definecolor{darkblue}{rgb}{0, 0.40, 0.75}

\usepackage{epigraph}
\usepackage{hyperref} 
\hypersetup{
    colorlinks=true,
    linkcolor=red,
    citecolor=cyan,
}
\usepackage{amssymb}
\usepackage{bbding}

\usepackage{booktabs}
\usepackage{color}
\usepackage[dvipsnames]{xcolor}

\usepackage{multirow}

\usepackage{caption}
\usepackage{subcaption}
\usepackage{amsmath}

\usepackage{amssymb}
\usepackage{pifont}

\usepackage[most]{tcolorbox}
\newtcolorbox{AIbox}[2][]{aibox,title=#2,#1}

\tcbset{
  aibox/.style={
    width=\linewidth,
    top=8pt,
    bottom=4pt,
    colback=blue!6!white,
    colframe=black,
    colbacktitle=black,
    enhanced,
    center,
    attach boxed title to top left={yshift=-0.1in,xshift=0.15in},
    boxed title style={boxrule=0pt,colframe=white,},
  }
}
\usepackage{enumitem} 

\begin{document}
%
\title{A Survey on Agentic Multimodal Large Language Models}
%
%
%
%

\author{
Huanjin Yao$^{\dagger}$, Ruifei Zhang$^{\dagger}$, Jiaxing Huang$\textsuperscript{\Envelope}$, Jingyi Zhang, Yibo Wang, Bo Fang, Ruolin Zhu, \\ Yongcheng Jing, Shunyu Liu, Guanbin Li, Dacheng Tao
\IEEEcompsocitemizethanks{
\IEEEcompsocthanksitem Huanjin Yao, Jiaxing Huang, Jingyi Zhang, Yibo Wang, Yongcheng Jing, Shunyu Liu, Dacheng Tao are with the Nanyang Technological University, Singapore.
\IEEEcompsocthanksitem Ruifei Zhang is with the Chinese University of Hong Kong, Shenzhen, China, and also with the Shenzhen Research Institute of Big Data, China.
\IEEEcompsocthanksitem Guanbin Li is with the Sun Yat-sen University, China.
\IEEEcompsocthanksitem Bo Fang is with the City University of Hong Kong, China.
\IEEEcompsocthanksitem Ruolin Zhu is with the Communication University of China, China.
\IEEEcompsocthanksitem $\dagger$ denotes equal contribution; $\textsuperscript{\Envelope}$ denotes corresponding author.
}
}

%
%

\markboth{Journal of \LaTeX\ Class Files, October~2025}%
{Shell \MakeLowercase{\textit{et al.}}: Bare Advanced Demo of IEEEtran.cls for IEEE Computer Society Journals}
%



\IEEEtitleabstractindextext{%
\begin{abstract}
With the recent emergence of revolutionary autonomous agentic systems, research community is witnessing a significant shift from traditional static, passive, and domain-specific AI agents toward more dynamic, proactive, and generalizable agentic AI.
Motivated by the growing interest in agentic AI and its potential trajectory toward AGI, 
we present a comprehensive survey on Agentic Multimodal Large Language Models (Agentic MLLMs).
In this survey, we explore the emerging paradigm of agentic MLLMs, delineating their conceptual foundations and distinguishing characteristics from conventional MLLM-based agents.
We establish a conceptual framework that organizes agentic MLLMs along three fundamental dimensions:
(i) \textbf{Agentic internal intelligence} functions as the system's commander, enabling accurate long-horizon planning through reasoning, reflection, and memory; 
(ii) \textbf{Agentic external tool invocation}, whereby models proactively use various external tools to extend their problem-solving capabilities beyond their intrinsic knowledge;
and (iii) \textbf{Agentic environment interaction} further situates models within virtual or physical environments, allowing them to take actions, adapt strategies, and sustain goal-directed behavior in dynamic real-world scenarios.
To further accelerate research in this area for the community, we compile open-source training frameworks, training and evaluation datasets for developing agentic MLLMs.
Finally, we review the downstream applications of agentic MLLMs and outline future research directions for this rapidly evolving field.
To continuously track developments in this rapidly evolving field, we will also actively update a public repository at \url{https://github.com/HJYao00/Awesome-Agentic-MLLMs}.

\end{abstract}

}

\maketitle

\IEEEdisplaynontitleabstractindextext

%
\IEEEpeerreviewmaketitle

\ifCLASSOPTIONcompsoc
\IEEEraisesectionheading{\section{Introduction}\label{sec:introduction}}
\else
\section{Introduction}
\label{sec:introduction}
\fi

\IEEEPARstart{M}{ulti-Modal} Large Language Models (MLLMs) have achieved remarkable progress in recent years, enabling AI systems to perceive, understand, reason, and generate across diverse modalities~\cite{openai2024GPT-4o, bai2025qwen2.5vl, team2025gemini, xu2025qwen2.5-omni, yao2024minicpm, wu2024nextgpt, zhan2024anygpt, claude-4}. With strong instruction-following ability and cross-modal generalization, MLLMs are capable of tackling a wide spectrum of tasks, making them increasingly valuable in both general applications and professional contexts~\cite{ye2025harnessing, wu2023gpt4vis, yang2023llm4drive, li2024manipllm, huang2025keeping, zhang2024vlm_survey}.
However, most traditional MLLMs still operate under a query–response paradigm, where static inputs produce single outputs. This paradigm is often inadequate for complex, dynamic real-world tasks, which require three essential capabilities: internal intelligence (e.g., reasoning~\cite{huang2025visionr1, zhang2025r1vl, wang2025mm_cot_survey, chen2025cot_rllm_survey}, reflection~\cite{yao2024mulberryempoweringmllmo1like, wan2025srpo}, and memory~\cite{xu2025a-mem, yan2025memoryr1}), external tool invocation (e.g., information searching~\cite{wu2025mmsearchr1, geng2025webwatcher}, code execution~\cite{feng2025retool, liu2025visualarft}, and visual processing~\cite{zheng2025deepeyes, lai2025mini-o3, su2025thinking_with_images_survey}), and environment interaction (e.g., virtual embodiment~\cite{luo2025guir1, lin2025showui} and physical embodiment~\cite{kim2024openvla, vlaathinker}).


To extend the capabilities of MLLMs beyond static query–response interactions, MLLM agents~\cite{xie2024large_mm_agent_survey, wang2024survey_llm_agent} have attracted increasing attention, which embeds MLLMs within structured workflows, enabling task decomposition, scenario-specific reasoning, and integration of external tools~\cite{cao2025phishagent, koh2024visualwebarena, deng2023mind2web, verma2024adaptagent, yao2022webshop, wang2024mobile-agent-v2}.
Despite their effectiveness, existing MLLM agents still suffer from several constraints: 1) Static workflow: they rely heavily on pre-defined and handcrafted workflows that are inflexible and cannot adapt to novel or dynamic situations; 2) Passive execution: they typically respond passively to instructions, without genuine intelligence to initiate plans, invoke tools, or proactively engage with environments; 3) Domain-specific application: most MLLM agents are tailored for a single task or domain, resulting in poor generalization and limited scalability across diverse domains or tasks.


\tikzstyle{my-box}=[
rectangle,
draw=hidden-black,
rounded corners,
text opacity=1,
minimum height=1.5em,
minimum width=5em,
inner sep=2pt,
align=center,
fill opacity=.5,
]
\tikzstyle{leaf}=[
my-box, 
minimum height=1.5em,
fill=yellow!32, 
text=black,
align=left,
font=\normalsize,
inner xsep=5pt,
inner ysep=4pt,
align=left,
text width=45em,
]
\tikzstyle{leaf2}=[
my-box, 
minimum height=1.5em,
fill=purple!27, 
text=black,
align=left,
font=\normalsize,
inner xsep=5pt,
inner ysep=4pt,
]
\tikzstyle{leaf3}=[
my-box, 
minimum height=1.5em,
fill=hidden-blue!57, 
text=black,
align=left,
font=\normalsize,
inner xsep=5pt,
inner ysep=4pt,
]

\tikzstyle{leaf4}=[
my-box, 
minimum height=1.5em,
fill=hidden-green!27, 
text=black,
align=left,
font=\normalsize,
inner xsep=5pt,
inner ysep=4pt,
]

\tikzstyle{leaf5}=[
my-box, 
minimum height=1.5em,
fill=hidden-orange!37, 
text=black,
align=left,
font=\normalsize,
inner xsep=5pt,
inner ysep=4pt,
]

\tikzstyle{leaf6}=[
my-box, 
minimum height=1.5em,
fill=violet!12, 
text=black,
align=left,
font=\normalsize,
inner xsep=5pt,
inner ysep=4pt,
]

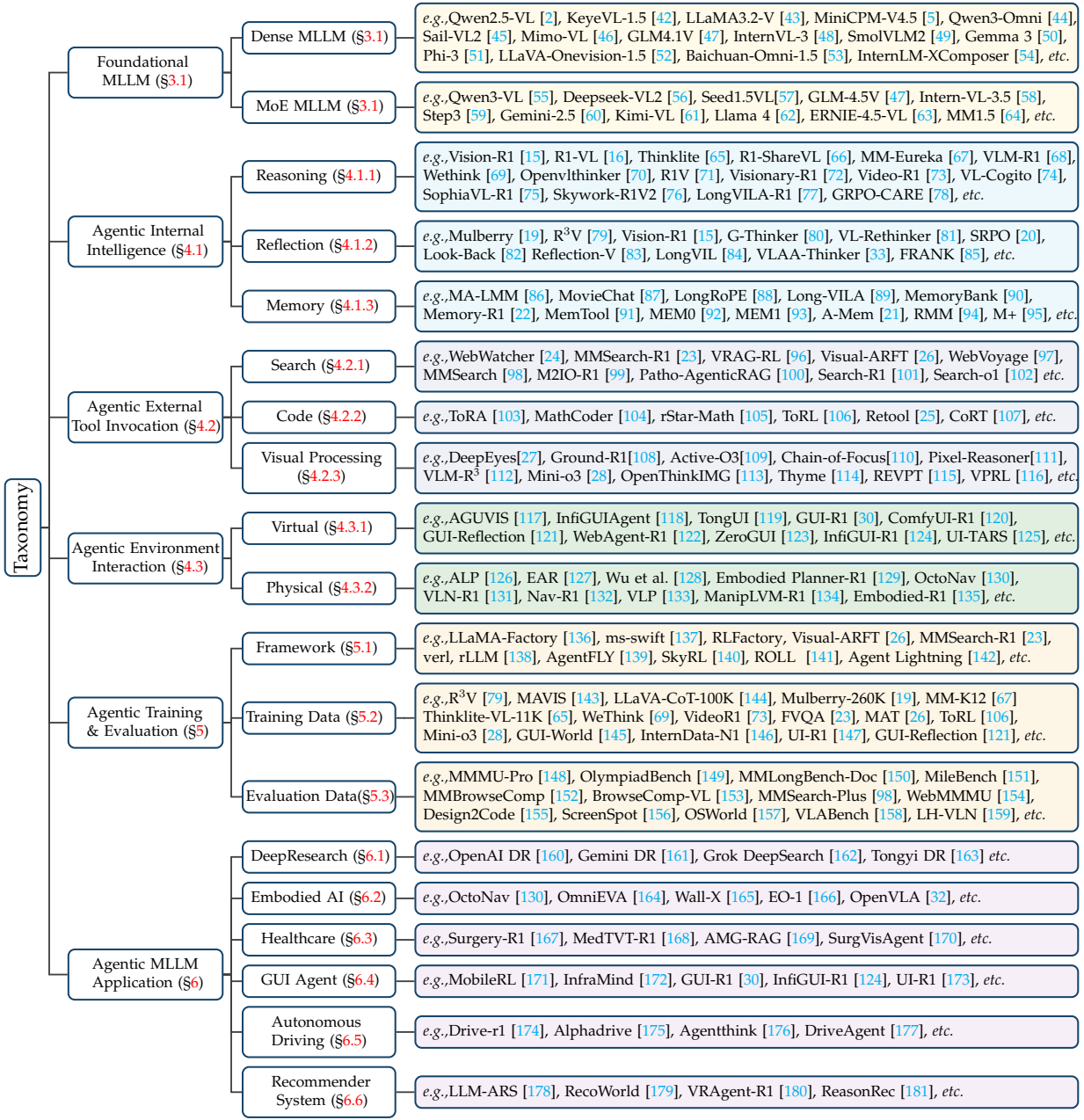
\begin{figure*}[ht]
\vspace{-2mm}
\centering
\resizebox{0.9\textwidth}{!}{
	\begin{forest}
		forked edges,
		for tree={
			grow=east,
			reversed=true,
			anchor=base west,
			parent anchor=east,
			child anchor=west,
			base=left,
			font=\large,
			rectangle,
			draw=hidden-black,
			rounded corners,
			align=left,
			minimum width=4em,
			edge+={darkgray, line width=1pt},
			s sep=5pt,
			inner xsep=2pt,
			inner ysep=4pt,
			line width=1.1pt,
			ver/.style={rotate=90, child anchor=north, parent anchor=south, anchor=center},
		},
		where level=1{text width=9.6em,font=\normalsize,}{},
        where level=2{text width=9.5em,font=\normalsize,}{},
        where level=3{text width=9.5em,font=\normalsize,}{},
        where level=4{text width=52em,font=\normalsize,}{},
[Taxonomy, ver
	[\ \ \ \ \ Foundational \\ \ \ \ \ \ \  MLLM~(\S\ref{sec: 3.1 foundational mllm}),
		[\ Dense MLLM~(\S\ref{sec: 3.1 foundational mllm})
                [\eg Qwen2.5-VL~\cite{bai2025qwen2.5vl}{,} KeyeVL-1.5~\cite{keyevl-1.5}{,} LLaMA3.2-V~\cite{llama3}{,} MiniCPM-V4.5~\cite{yao2024minicpm}{,} Qwen3-Omni~\cite{xu2025qwen3-omni}{,} \\
                Sail-VL2~\cite{yin2025sailvl2technicalreport}{,} Mimo-VL~\cite{xiaomi2025mimo-vl}{,} GLM4.1V~\cite{glm4.5v}{,} InternVL-3~\cite{zhu2025internvl3}{,} SmolVLM2~\cite{marafioti2025smolvlm}{,} Gemma 3~\cite{team2025gemma3}{,} \\
                Phi-3~\cite{abdin2024phi4}{,} LLaVA-Onevision-1.5~\cite{LLaVA-OneVision-1.5}{,} Baichuan-Omni-1.5~\cite{li2025baichuan-omni-1.5}{,} InternLM-XComposer~\cite{dong2024internlm-xcomposer2}{,} \textit{etc.}, leaf, text width=42em]
            ]
		[\ \ MoE MLLM~(\S\ref{sec: 3.1 foundational mllm})
                [\eg Qwen3-VL~\cite{qwen3-vl}{,} Deepseek-VL2~\cite{wu2024deepseek-vl2}{,} Seed1.5VL\cite{guo2025seed1.5-vl}{,} GLM-4.5V~\cite{glm4.5v}{,} Intern-VL-3.5~\cite{wang2025internvl3.5}{,} \\ Step3~\cite{step3system}{,}  Gemini-2.5~\cite{comanici2025gemini-2.5}{,} Kimi-VL~\cite{team2025kimivl}{,} Llama 4~\cite{Llama4}{,} ERNIE-4.5-VL~\cite{ERNIE-4.5}{,} MM1.5~\cite{zhang2024mm1.5}{,} \textit{etc.}
                , leaf, text width=42em]
		]
	]
	[\ \ \ \ Agentic Internal \\ \ \ \  Intelligence~(\S\ref{Sec: 4.1 internal self-planning}),
		[\ \ Reasoning~(\S\ref{sec: 4.1.1 agentic reasoning})
                [\eg Vision-R1~\cite{huang2025visionr1}{,} R1-VL~\cite{zhang2025r1vl}{,} Thinklite~\cite{wang2025thinklitevl}{,} R1-ShareVL~\cite{yao2025r1sharevl}{,} MM-Eureka~\cite{meng2025mmekura}{,} VLM-R1~\cite{shen2025vlm-r1}{,}  \\
                    Wethink~\cite{yang2025wethink}{,} Openvlthinker~\cite{deng2025openvlthinkerearlyexplorationcomplex}{,} R1V~\cite{chen2025r1v}{,} Visionary-R1~\cite{xia2025visionary-r1}{,} Video-R1~\cite{feng2025videor1}{,} VL-Cogito~\cite{yuan2025vlcogito}{,} \\
                    SophiaVL-R1~\cite{fan2025sophiavl}{,} Skywork-R1V2~\cite{wang2025skyworkr1v2}{,} LongVILA-R1~\cite{longvila-r1}{,} GRPO-CARE~\cite{chen2025grpocare}{,}
                    \textit{etc.}
				, leaf3, text width=42em
                ]
		]
		[\ \ Reflection~(\S\ref{sec: 4.1.2 agentic reflection})
                [\eg Mulberry~\cite{yao2024mulberryempoweringmllmo1like}{,} R$^3$V~\cite{r3v_reflection}{,} Vision-R1~\cite{huang2025visionr1}{,} G-Thinker~\cite{zhan2025gthinker}{,} VL-Rethinker~\cite{wang2025vlrethinker}{,} SRPO~\cite{wan2025srpo}{,} \\
                Look-Back~\cite{yang2025Look-Back} Reflection-V~\cite{jian2025reflection-v}{,} LongVIL~\cite{chen2025longvil}{,} VLAA-Thinker~\cite{vlaathinker}{,} FRANK~\cite{frank}{,} \textit{etc.}
				, leaf3, text width=42em
			]
		]
            [\ \ \ \ Memory~(\S\ref{sec: 4.1.3 agentic memory})
                [\eg MA-LMM~\cite{he2024malmm}{,} MovieChat~\cite{song2024moviechat}{,} LongRoPE~\cite{ding2024longrope}{,} Long-VILA~\cite{chen2024longvila}{,}  MemoryBank~\cite{zhong2023memorybank}{,} \\
                Memory-R1~\cite{yan2025memoryr1}{,} MemTool~\cite{lumer2025memtool}{,} MEM0~\cite{chhikara2025mem0}{,} MEM1~\cite{zhou2025mem1}{,} A-Mem~\cite{xu2025a-mem}{,} RMM~\cite{tan2025prospect_rmm}{,} M+~\cite{wang2025m+}{,} \textit{etc.}
				, leaf3, text width=42em
			]
		]
	]
	[\ \ \ Agentic External \\ Tool Invocation~(\S\ref{Sec: 4.2 external tool}),
    [\ \ \ \ \  Search~(\S\ref{sec: 4.2.1 agentic search})
			[\eg WebWatcher~\cite{geng2025webwatcher}{,} MMSearch-R1~\cite{wu2025mmsearchr1}{,} VRAG-RL~\cite{wang2025vrag-rl}{,} Visual-ARFT~\cite{liu2025visualarft}{,} WebVoyage~\cite{he2024webvoyager}{,} \\
            MMSearch~\cite{tao2025mmsearch}{,} M2IO-R1~\cite{xiao2025m2io-r1}{,} Patho-AgenticRAG~\cite{zhang2025Patho-AgenticRAG}{,} Search-R1~\cite{jin2025searchr1}{,} Search-o1~\cite{li2025search-o1} \textit{etc.}, leaf2, text width=42em
		]
	]
	[\ \ \ \ \ \ Code~(\S\ref{sec: 4.2.2 agentic code})
		[\eg ToRA~\cite{gou2023tora}{,} MathCoder~\cite{wang2023mathcoder}{,} rStar-Math~\cite{guan2025rstar}{,} ToRL~\cite{li2025torl}{,} Retool~\cite{feng2025retool}{,} CoRT~\cite{li2025cort}{,}
        \textit{etc.}
			, leaf2, text width=42em
		]
    ]
	[ \ \ \ Visual Processing \\ ~~~~~~~~~~(\S\ref{sec: 4.2.3 agentic data process})
			[\eg DeepEyes\cite{zheng2025deepeyes}{,} Ground-R1\cite{cao2025ground}{,}  Active-O3\cite{zhu2025active}{,} Chain-of-Focus\cite{zhang2025chain}{,} 
            Pixel-Reasoner\cite{su2025pixel}{,} \\         
            VLM-R\textsuperscript{3}~\cite{jiang2025vlm}{,} Mini-o3~\cite{lai2025mini-o3}{,} OpenThinkIMG~\cite{su2025openthinkimg}{,}
            Thyme~\cite{zhang2025thyme}{,}          
            REVPT~\cite{zhou2025reinforced}{,}
            VPRL~\cite{xu2025visual}{,}
            \textit{etc.}
			, leaf2, text width=42em
		]
	]
    ]
    [Agentic Environment \\ \ \ \ \  Interaction~(\S\ref{Sec: 4.3 environment interaction}),
		[\ \ \ \ \ Virtual~(\S\ref{sec: 4.3.1 agentic Virtual Embodiment})
				[\eg AGUVIS~\cite{xu2024aguvis}{,}          InfiGUIAgent~\cite{liu2025infiguiagent}{,} TongUI~\cite{zhang2025tongui}{,} GUI-R1~\cite{luo2025guir1}{,} ComfyUI-R1~\cite{xu2025comfyui}{,}  \\ 
                GUI-Reflection~\cite{wu2025gui-reflection}{,}
                WebAgent-R1~\cite{wei2025webagent}{,}
                ZeroGUI~\cite{yang2025zerogui}{,}
                InfiGUI-R1~\cite{liu2025infiguir1}{,} UI-TARS~\cite{qin2025ui}{,} \textit{etc.}
				, leaf4, text width=42em
			]
		]
		[\ \ \ \ Physical~(\S\ref{sec: 4.3.1 agentic Physical Embodiment})
				[\eg ALP~\cite{liang2023alp}{,} EAR~\cite{fan2024evidential}{,} Wu et al.~\cite{wu2025reinforced}{,} Embodied Planner-R1~\cite{fei2025unleashing}{,} OctoNav~\cite{gao2025octonav}{,} \\ VLN-R1~\cite{qi2025vln}{,}
                Nav-R1~\cite{liu2025navr1}{,}  VLP~\cite{liu2025vlp}{,} ManipLVM-R1~\cite{song2025maniplvm}{,} Embodied-R1~\cite{yuan2025embodied}{,} \textit{etc.}
				, leaf4, text width=42em
			]
		]
	]
        [\ \ \ Agentic Training \\ \ \ \ \& Evaluation~(\S\ref{sec: 5 training and evaluation}),
            [\ \  Framework~(\S\ref{sec: 5.1 framework})
                [\eg LLaMA-Factory~\cite{zheng2024llamafactory}{,} ms-swift~\cite{zhao2024swift}{,} 
                     RLFactory{,} Visual-ARFT~\cite{liu2025visualarft}{,} MMSearch-R1~\cite{wu2025mmsearchr1}{,} \\
                     verl{,} rLLM~\cite{rllm2025}{,} AgentFLY~\cite{wang2025agentfly}{,} SkyRL~\cite{cao2025skyrl}{,} ROLL ~\cite{wang2025reinforcement}{,} Agent Lightning~\cite{luo2025agentlightningtrainai}{,} \textit{etc.}
                , leaf5, text width=42em]
            ]
            [Training Data~(\S\ref{sec: 5.2 training dataset})
                [\eg R$^3$V~\cite{r3v_reflection}{,} MAVIS~\cite{zhang2024mavis}{,} LLaVA-CoT-100K~\cite{xu2025llavacotletvisionlanguage}{,} Mulberry-260K~\cite{yao2024mulberryempoweringmllmo1like}{,} MM-K12~\cite{meng2025mmekura}  \\
                Thinklite-VL-11K~\cite{wang2025thinklitevl}{,} WeThink~\cite{yang2025wethink}{,} VideoR1~\cite{feng2025videor1}{,} FVQA~\cite{wu2025mmsearchr1}{,} MAT~\cite{liu2025visualarft}{,} ToRL~\cite{li2025torl}{,} \\ Mini-o3~\cite{lai2025mini-o3}{,}
                GUI-World~\cite{chen2024gui-world}{,} InternData-N1~\cite{interndata_n1}{,} UI-R1~\cite{GUI-Critic-R1}{,} GUI-Reflection~\cite{wu2025gui-reflection}{,} \textit{etc.}
                , leaf5, text width=42em]
            ]
            [Evaluation Data(\S\ref{sec: 5.3 evaluation dataset})
            [\eg MMMU-Pro~\cite{yue2024mmmu-pro}{,}  OlympiadBench~\cite{he2024olympiadbench}{,} MMLongBench-Doc~\cite{ma2024mmlongbench}{,} MileBench~\cite{song2024milebench}{,} \\
            MMBrowseComp~\cite{li2025mm-BrowseComp}{,} BrowseComp-VL~\cite{wei2025browsecomp}{,} MMSearch-Plus~\cite{tao2025mmsearch}{,} WebMMMU~\cite{awal2025webmmu}{,} \\ Design2Code~\cite{si2024design2code}{,} ScreenSpot~\cite{liu2025screens}{,} OSWorld~\cite{xie2024osworld}{,} VLABench~\cite{zhang2024vlabench}{,} LH-VLN~\cite{lh-vln}{,} \textit{etc.}
            , leaf5, text width=42em]
            ]
	]
        [\ \ \ \ Agentic MLLM \\ \ \ \ \ Application~(\S\ref{sec: 6 application}),
            [\ DeepResearch~(\S\ref{sec: 6.1 deep research})
                [\eg OpenAI DR~\cite{openai2025deepresearch}{,} Gemini DR~\cite{google2025deepresearch}{,} Grok DeepSearch~\cite{xai2025deepsearch}{,} Tongyi DR~\cite{tongyidr} \textit{etc.}
                , leaf6, text width=42em]
            ]
            [\ Embodied AI~(\S\ref{sec: 6.2 Embodied AI})  \ \ \ \ 
                [\eg OctoNav~\cite{gao2025octonav}{,} OmniEVA~\cite{liu2025omnieva}{,} Wall-X~\cite{wall-x}{,} EO-1~\cite{qu2025embodiedonevision}{,} OpenVLA~\cite{kim2024openvla}{,} \textit{etc.}
                , leaf6, text width=42em]
            ]
            [\ \ \ Healthcare~(\S\ref{sec: 6.3 healthcare})
                [\eg Surgery-R1~\cite{hao2025surgery}{,} MedTVT-R1~\cite{zhang2025medtvt}{,} AMG-RAG~\cite{amg-rag}{,} SurgVisAgent~\cite{lei2025surgvisagent}{,} \textit{etc.}
                , leaf6, text width=42em]
            ]
            [\ \ \ GUI Agent~(\S\ref{sec: 6.4 GUI_agents}) \ \ \ \ 
                [\eg MobileRL~\cite{xu2025mobilerl}{,} InfraMind~\cite{lin2025inframind}{,} GUI-R1~\cite{luo2025guir1}{,} InfiGUI-R1~\cite{liu2025infiguir1}{,} UI-R1~\cite{lu2025uir1}{,} \textit{etc.}
                , leaf6, text width=42em]
            ]
            [\ \ \ \ \ Autonomous  \\ \ \ \ \ \  Driving~(\S\ref{sec: 6.5 autonomous_driving})
                [\eg Drive-r1~\cite{li2025drive}{,} Alphadrive~\cite{jiang2025alphadrive}{,} Agentthink~\cite{qian2025agentthink}{,} DriveAgent~\cite{hou2025driveagent}{,} \textit{etc.}
                , leaf6, text width=42em]
            ]
            [\ \ \ \ \  Recommender \\ \ \ \ \ \ \  System~(\S\ref{sec: 6.6 Recommender System})
                [\eg LLM-ARS~\cite{LLM-ARS}{,} RecoWorld~\cite{liu2025recoworld}{,} VRAgent-R1~\cite{chen2025vragent-r1}{,} ReasonRec~\cite{zhang2025reasonrec}{,} \textit{etc.}
                , leaf6, text width=42em]
            ]
	   ]
]
	\end{forest}
}
\caption{The primary organizational structure of the survey and key works illustrating progress in each direction.}
\label{fig: taxonomy}

\end{figure*}

Recent advances in reasoning-enhanced MLLMs~\cite{jaech2024openai-o1, glm4.5v, team2025kimivl, ke2025survey_llm_reasoning} and reinforcement learning (RL)~\cite{guo2025deepseek-r1, shao2024deepseekmath, zhang2025landscape_agentic_rl_llm_sruvey, zhang2025survey_rl_lrm} have driven a paradigm shift from workflow-bound MLLM agents toward agentic MLLMs.
Unlike traditional agents, agentic MLLMs~\cite{geng2025webwatcher, tongyidr, openai2025deepresearch,2025mirothinker, google2025deepresearch, perplexity2025deepresearch} are framed as autonomous decision-makers, which possess built-in agentic capabilities, i.e., the autonomy to reason, reflect, memory, use tools, and interact with environments.
To this end, agentic MLLMs offer several key advantages:
(1) First, agentic MLLMs can dynamically adjust their strategies and workflows based on previous planning, current state, and anticipated environmental interactions rather than relying on static, pre-defined and handcrafted procedures.
(2) Second, agentic MLLMs plan and execute actions proactively, autonomously initiating plans, invoking tools when needed, and reflecting on intermediate outcomes to refine subsequent steps.
(3) Third, agentic MLLMs can operate across diverse tasks and environments, enabling general-purpose modeling and learning, instead of being restricted to narrow, domain-specific applications.
This transition marks not only stronger planning capabilities, but also genuine intelligence: the ability to generate plans adaptively, invoke tools proactively, and engage effectively with dynamic environments.

Despite the growing attention on advancing agentic MLLMs, the research community still lacks a comprehensive survey that can help organize current progress, identify key challenges, and highlight promising directions in this rapidly evolving field.
To fill this gap, we present a systematic review of agentic MLLMs over three major components including agentic internal intelligence, agentic external tool invocation, and agentic environment interaction.
We conduct the survey from different perspectives including discussion, foundations, technical approaches, training \& evaluation resources, and future research directions.
We expect this survey to provide a thorough overview of current achievements and to outline the pathways for further progress in this rapidly evolving and promising area.

In summary, the main contributions of this work are threefold.
1) it presents a systematic review of the development of agentic MLLMs, categorizing existing studies according to different tasks. To the best of our knowledge, this is the first survey in this field, offering an overarching view and thorough classification.
2) it studies the up-to-date progress of agentic MLLMs, including methodological advances as well as training and evaluation resources, with corresponding links provided for ease of reference.
3) it shares several research challenges and potential research directions that could be pursued in agentic MLLMs.

\begin{figure*}[ht]
\centering
\includegraphics[width=0.99\linewidth]{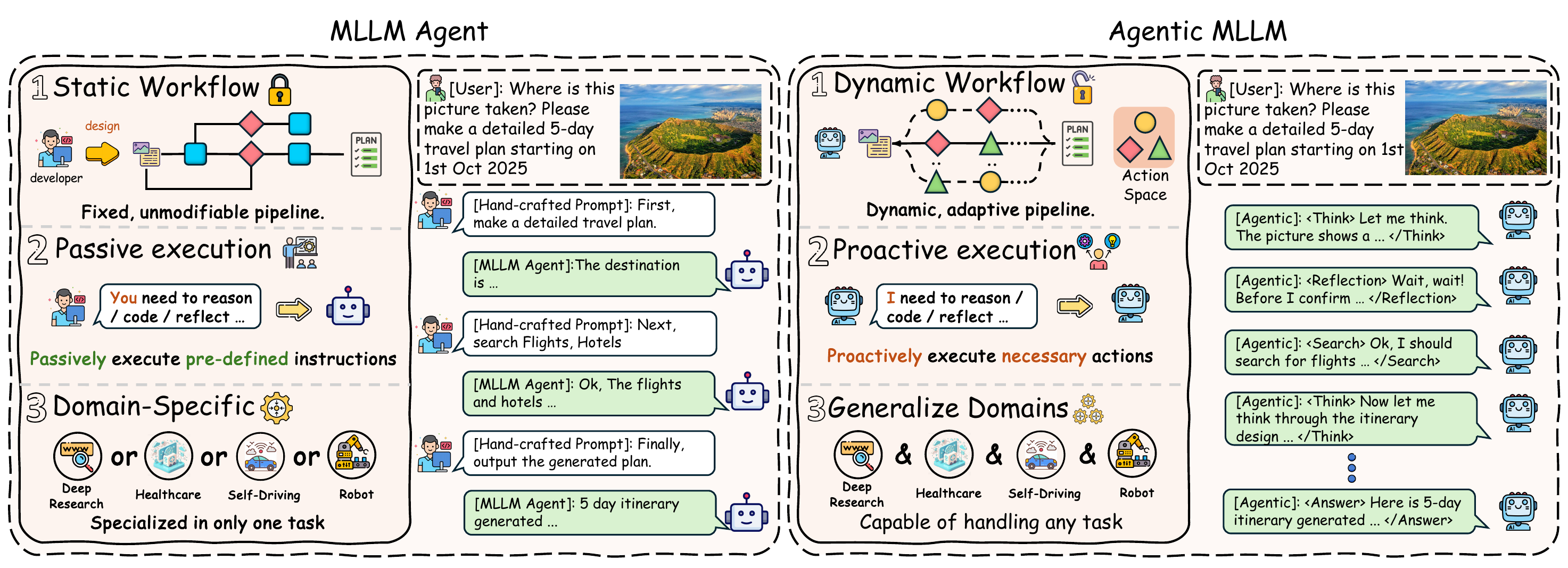}
\caption{
The key differences between Agentic MLLMs and MLLM Agents lie in three defining characteristics of Agentic MLLMs: a dynamic and adaptive workflow, proactive execution of actions, and strong generalization across domains.
}
\label{fig:2 agent vs agentic}
\vskip -0.1in
\end{figure*}

To this end, our survey is organized according to the taxonomy illustrated in Figure~\ref{fig: taxonomy}. The rest of this survey is organized as follows.
Section~\ref{sec: 2} presents the discussion of MLLM agents and agentic MLLMs.
We then introduce the foundational concepts of agentic MLLMs in Section~\ref{sec: 3}, encompassing foundational MLLMs, agentic action space, agentic MLLM training and evaluation.
Section~\ref{sec: 4 agentic mllm} reviews and categorizes existing agentic MLLM studies, including agentic internal intelligence, agentic external tool invocation, and agentic environment interaction.
Section~\ref{sec: 5 training and evaluation} presents the widely-used training frameworks, training and evaluation datasets for agentic MLLMs.
Section~\ref{sec: 6 application} introduces the applications of agentic MLLMs, such as DeepResearch, Embodied AI, Healthcare, GUI Agents, Autonomous Driving, and Recommender System.
Finally, we share several promising agentic MLLMs research directions in Section \ref{sec: 7 challenges and future directions}.





\section{Discussion of MLLM Agent and Agentic MLLM}
\label{sec: 2}

This section formalizes the key distinctions between agentic MLLMs and conventional MLLM agents, emphasizing the dynamic workflows, proactive action execution, and cross-domain generalization capabilities of agentic MLLMs, as illustrated in Figure~\ref{fig:2 agent vs agentic}.

\subsection{Overview of MLLM Agents}
MLLM agents~\cite{xie2024large_mm_agent_survey, li2024autoflow, wang2024videoagent, jiang2024mmsearch, he2024webvoyager} are typically defined by a static workflow that is meticulously pre-designed and implemented by developers, adhering to a divide-and-conquer principle~\cite{gu2025agentgroupchat-v2}.
In this paradigm, a complex task is decomposed in a flowchart-like structure into a series of smaller subtasks, with the MLLM assigned different roles through carefully crafted prompts at each stage.
Then, these role-specific instances of the MLLM execute their respective instructions within an orchestrated workflow, where intermediate outputs are cascaded downstream to subsequent stages. Ultimately, the process yields a complete solution in a modular manner, which can be formalized as follows:
\begin{align}
\text{Agent}_{\text{MLLM}} = f_T \circ f_{T-1} \circ \cdots \circ f_1(x_1)
\end{align}
\begin{align}
f_i(x_i) = \text{MLLM}(p_i, x_i), ~ x_{i+1} = f_i(x_i).
\end{align}
where $p_i$ represents the manually crafted prompt at stage $i$, $f_i$ denotes the responses of the MLLM conditioned on prompt $p_i$, and $x_{i+1}$ is the sequential multimodal input passed forward from the previous stage. After all subtasks are completed in sequence, the overall process of $\text{Agent}_{\text{MLLM}}$ ultimately produces the final output.

Overall, MLLM agents position the MLLM as a task executor capable of accomplishing complex objectives through systematic decomposition into subtasks.
However, their intrinsic design is bound to a static and fixed workflow, where roles are assigned exclusively through predefined prompts.
This constraint results in static planning, passive action execution, and domain-specific limitations, as illustrated in Figure~\ref{fig:2 agent vs agentic}, hindering adaptability and generalizability.


\subsection{Overview of Agentic MLLMs}
In contrast, agentic MLLMs treat task-solving as an autonomous decision-making process, in which the model independently selects actions at each step in response to contextual features and evolving environmental states.
As illustrated in Figure~\ref{fig:2 agent vs agentic}, we highlight three fundamental distinctions between MLLM agents and agentic MLLMs, which are elaborated in the following subsections.

\subsubsection{Dynamic Workflow}
As shown in Figure~\ref{fig:2 agent vs agentic}, traditional MLLM agents rely on a static and unmodifiable pipeline pre-designed by developers to solve the problem.
In contrast, agentic MLLMs dynamically select appropriate strategies in response to the evolving state, enabling an adaptive problem-solving process and breaking free from fixed execution patterns.
This dynamic workflow and its underlying state transitions can be represented at each step as:
\begin{align}
s_{t+1} &= \delta\!\left(s_t, \; a_t\right),
\end{align}
where $s_t$ denotes the current state, $a_t$ is the action chosen by the MLLMs, and $\delta$ represents the state transition function.

        

\subsubsection{Proactive Action Execution}
As illustrated in Figure~\ref{fig:2 agent vs agentic}, conventional MLLM agents passively execute actions at each stage according to pre-defined instructions designed by developers.
In contrast, agentic MLLMs adopt a proactive paradigm in which actions are autonomously selected at every step based on the current state. This shift moves the model from simply following instructions to actively planning about ``what action should be taken next,'' thereby substantially improving its capacity for context-sensitive decision-making. Formally, proactive action execution can be expressed as:
\begin{align}
a_t \sim \pi(a \mid s_t),
\end{align}
where $a_t$ denotes the action chosen under the current state $s_t$ according to the policy $\pi$.


\subsubsection{Generalization Across Domains}
As illustrated in Figure~\ref{fig:2 agent vs agentic}, traditional MLLM agents require developers to design bespoke pipelines and prompts for each task, rendering them domain-specific and limiting their ability to generalize to new scenarios.
In contrast, agentic MLLMs can adapt their workflows across evolving environments by adaptively planning and executing the actions required.
This flexibility enables them to operate in diverse contexts and to effectively solve tasks spanning multiple domains.
Formally, such Generalization can be formulated as a policy optimization objective that maximizes the expected cumulative reward:
\begin{align}
\pi^* &= \arg\max_{\pi} \; \mathbb{E}_{(x) \sim \mathcal{D}} 
\Big[ \sum_{t=0}^{T} \gamma^t \, r(s_t,a_t; x) \Big],
\end{align}
where $\mathcal{D}$ denotes the distribution of tasks and environments, $s_t$ is the state at step $t$, $a_t$ is the action sampled from policy $\pi$, $r(\cdot)$ is the reward function that drives generalization across domains, and $\gamma$ is the discount factor controlling the relative importance of long-term versus short-term rewards.

In summary, agentic MLLMs reconceptualize task-solving within the formalism of an action-oriented markov decision process. 
Rather than relying on static, hand-crafted pipelines, they are modeled as adaptive policies that interact with action space and environment, continually updating internal states and proactively making context-sensitive decisions. 
This formulation highlights their ability to autonomously plan, act, and generalize across diverse tasks and domains.
\begin{align}
\text{Agentic}_{\text{MLLM}} &= \pi^*(x, \mathcal{A}, \mathcal{E}),
\end{align}
\noindent where $x$ denotes the input, $\mathcal{A}$ the action space, and $\mathcal{E}$ the environment. 
Here, $\pi^*$ represents the optimal policy that governs adaptive decision-making across states, actions, and environmental dynamics.

\section{Foundations of Agentic MLLMs}
\label{sec: 3}
In this section, we introduce the preliminaries of agentic MLLMs covering:
(1) agentic foundational MLLMs, which serve as the base models for agentic systems;
(2) agentic action space, which defines how actions are formally specified and subsequently executed by the model;
(3) agentic continual pre-training, which equips MLLMs with broader agentic general knowledge;
(4) agentic supervised fine-tuning, which uses curated high-quality multi-turn trajectories to provide a cold start for RL;
(5) agentic reinforcement learning, which incentivizes agentic behavior through exploration and feedback; 
(6) agentic evaluation, which assesses model at the process level or the outcome level.

\subsection{Agentic Foundational MLLMs}
\label{sec: 3.1 foundational mllm}
Early foundational MLLMs~\cite{li2023blip2, alayrac2022flamingo, luo2023valley, llava, llava1.5, gpt4o, bai2025qwen2.5vl, li2024mini, lin2023video-llava} demonstrated the ability to jointly process and align images and text, achieving strong performance on a wide range of visual understanding tasks such as visual question answering~\cite{li2024llavaonevision, tong2024cambrian-1, idefics}, optical character recognition~\cite{ye2023ureader, chen2025ocean}, and table understanding~\cite{mplug-docowl-1.5, zhang2024tinychart}. These advances mark a transformative milestone in the multimodal field, positioning MLLMs as versatile multimodal systems capable of tackling a broad spectrum of tasks.

From an architectural perspective~\cite{feng2025dive, wang2024scaling_laws_dense_moe}, foundational MLLMs can be broadly categorized into two types: \textbf{dense MLLMs}, which activate all parameters during inference, and \textbf{Mixture-of-Experts (MoE) MLLMs}, which incorporate multiple experts but activate them sparsely. With the advent of agentic MLLMs, there has been an increasing trend toward MoE architectures, as multiple experts offer better support for adaptive reasoning and dynamic tool invocation. In the following, we review recent progress in dense MLLMs and MoE MLLMs separately.

\textbf{Dense MLLMs}:
Dense models are the classic architecture for MLLMs~\cite{bai2025qwen2.5vl, yin2025sailvl2technicalreport,wu2025valley2, llama3, llava1.5, zhu2025internvl3}, in which a single expert (i.e., a Feed-Forward Network) is employed and all parameters are activated for every input token. The forward computation is given by:
\begin{align}
&h^{(l+1)} = f\big(W^{(l)} h^{(l)} + b^{(l)}\big) \\
&f(h) = \sigma\big(W_{2} \, \sigma(W_{1} h + b_{1}) + b_{2}\big)
\end{align}
where $h^{(l)}$ denotes the input at layer $l$, $W^{(l)}$ and $b^{(l)}$ are the corresponding weight matrices and bias terms, and $f(\cdot)$ represents the feed-forward transformation with non-linear activation $\sigma(\cdot)$.
Each forward pass utilizes the full set of weights across all layers.
This design is straightforward, making optimization and deployment easy and stable.

Early pioneering open-source works on dense MLLMs, such as LLaVA~\cite{llava}, Flamingo~\cite{alayrac2022flamingo}, and BLIP-2~\cite{li2023blip2}, laid the foundation for multimodal understanding.
More recently, a series of follow-up studies, such as Qwen2.5-VL~\cite{bai2025qwen2.5vl}, MiniCPM-V 4.5~\cite{yao2024minicpm}, MiMoVL~\cite{xiaomi2025mimo-vl}, and Keye-VL-1.5~\cite{keyevl-1.5}, have further advanced the general multimodal understanding capabilities of dense MLLMs by leveraging more powerful language models~\cite{yang2025qwen3, zeng2025glm4.5, llama3}, scaling up training data~\cite{idefics, tong2024cambrian-1, li2024llavaonevision, FineVision}, and adopting improved optimization techniques~\cite{internvl-mpo, zhang2024amp, wang2024mmpr, NRCA_ICML25}.

\textbf{MoE MLLMs}: 
To expand model capacity (i.e., model size) without incurring prohibitive computational costs, many foundational MLLMs adopt a Mixture-of-Experts (MoE) architecture~\cite{lin2024moe-llava, li2023pace, chen2024eve, li2025uni-moe, huai2025cl-moe}.
In this design, a sparse activation mechanism ensures that only a small subset of experts is selected for each token.
A trainable gating network dynamically determines the routing of inputs to experts, allowing the model to scale to billions or even trillions of parameters while keeping the per-token computational cost comparable to that of smaller dense architectures. 
Such a mechanism enables specialization among experts, improves efficiency during inference, and facilitates handling of diverse multimodal tasks. 
Formally, the forward computation can be expressed as:
\begin{align}
&h^{(l+1)} = \sum_{i=1}^{K} g_i(x) \, f_i(h^{(l)}) \\
&f_i(h) = \sigma\big(W_{2,i} \, \sigma(W_{1,i} h + b_{1,i}) + b_{2,i}\big) \\
&g_i(x) = \frac{\exp(w_i^\top x)}{\sum_{j=1}^{K} \exp(w_j^\top x)}
\end{align}
\noindent where $f_i(\cdot)$ denotes the $i$-th Feed-Forward Network expert, $g_i(x)$ is the gating function that assigns routing weights to each expert, and $\sigma(\cdot)$ is a non-linear activation function. 
In practice, only the top-$k$ experts with the highest gating weights are activated, ensuring sparse computation and improved efficiency.
This makes one large model act like many specialized ones, better supporting varying levels of reasoning effort~\cite{agarwal2025gptoss, two_expert_for_think} and diverse agentic behaviors through adaptive expert selection~\cite{agarwal2025gptoss, zeng2025glm4.5}.

Recent work, such as Deepseek-VL2~\cite{wu2024deepseek-vl2}, which adopts DeepSeekMoE~\cite{dai2024deepseekmoe} as its language model, has demonstrated strong visual capabilities. 
GLM-4.5V~\cite{glm4.5v} contains a total of 106B parameters, but only 12B are activated during inference, substantially enhancing its reasoning capabilities.
Other studies including Kimi-VL~\cite{team2025kimivl}, Gemini-2.5~\cite{comanici2025gemini-2.5}, and Step-3~\cite{step3system} have also leveraged MoE architectures to further enhance their performance on complex tasks.
Moreover, GPT-oss~\cite{agarwal2025gptoss}, an MoE-based LLM, supports varying levels of reasoning effort and possesses native agentic capabilities. Building on this foundation, Intern-VL-3.5~\cite{wang2025internvl3.5} extends GPT-oss into the vision-language domain.

\begin{figure*}[ht]
\centering
\includegraphics[width=0.95\linewidth]{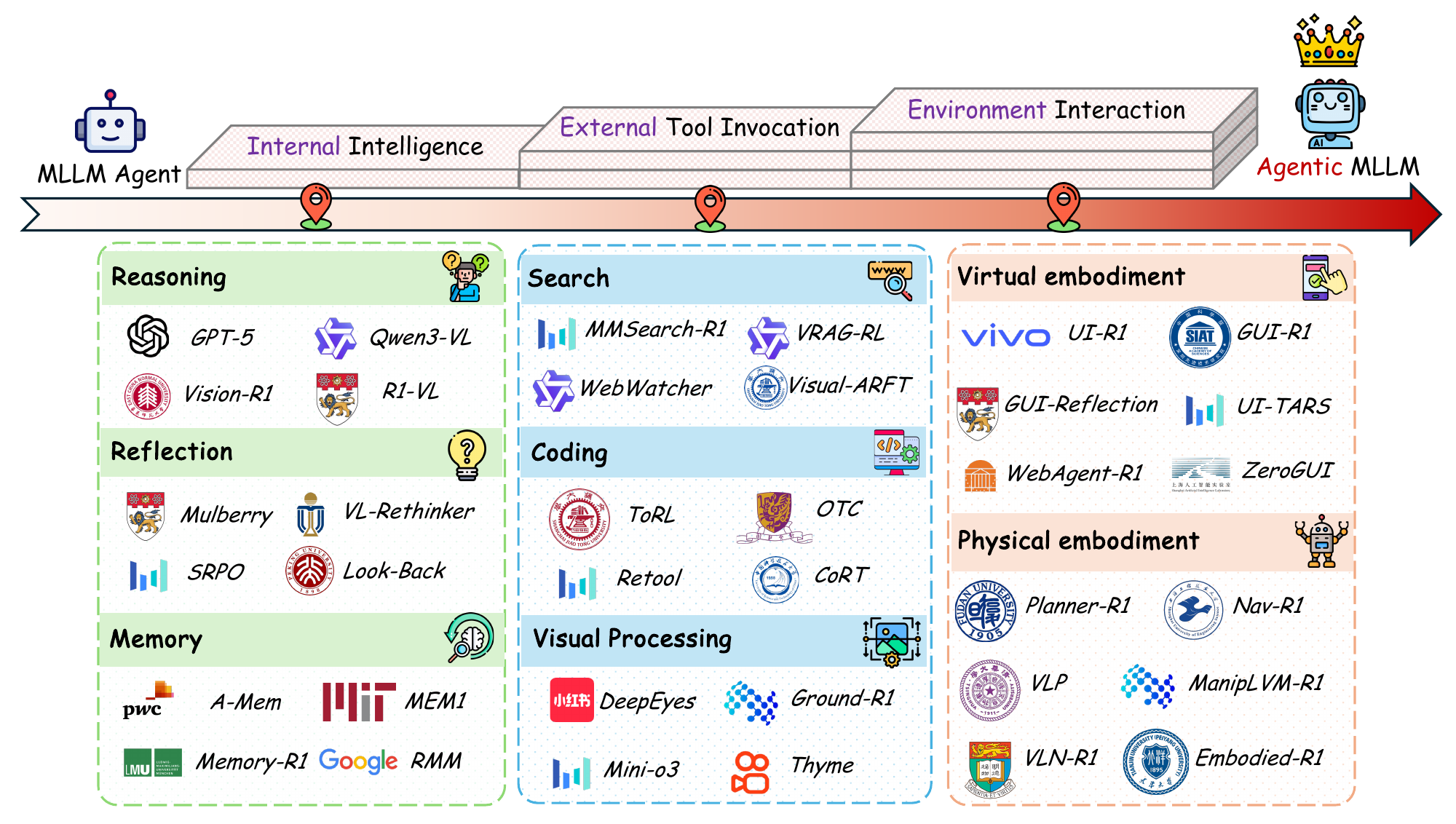}
\caption{The capability evolution from MLLM Agents to Agentic MLLMs: taxonomy and representative works across internal intelligence, external tool usage, and environmental interaction.}
\label{fig:3 overview of survey}
\vskip -0.1in
\end{figure*}

\subsection{Agentic Action Space}
Leveraging natural language as an interaction medium, MLLMs ground the definition of the action space in linguistic form, enabling the flexible and interpretable specification and execution of diverse actions.
Such actions may include reasoning, reflection, memory, various tools invocation, virtual and physical environment interaction, etc.
We summarize two approaches for embedding different actions into MLLMs, which are introduced in detail below.

\begin{itemize}
    \item \textbf{Specific Tokens.} Some studies~\cite{jin2025searchr1, liu2025visualarft, wu2025mmsearchr1} define different actions using distinct special tokens, such as \texttt{<action\_1> ... </action\_1>} and \texttt{<action\_2> ... </action\_2>}, where the content between the action tokens specifies the corresponding operation.
    \item \textbf{Unified Tokens}. Other studies~\cite{geng2025webwatcher, agarwal2025gptoss} adopt a more unified approach by invoking actions with a generic \texttt{<action>...</action>} token, within which a JSON-like structure specifies the tool to be called. For example: \texttt{<action>\{`action\_name': `action\_1', content': `...'\}</action>}.
\end{itemize}

At each state, agentic MLLMs reason over possible actions and select the best one that optimizes task completion, empowering autonomous decision-making and problem-solving capabilities that go far beyond a simple query–response chatbot.

\subsection{Agentic Contiunal Pre-training}
Agentic Continual Pre-training (Agentic CPT)~\cite{su2025agenticCPT} equips MLLMs with the ability to continually integrate new, up-to-date knowledge from diverse domains while enhancing their planning and tool-use capabilities, all without forgetting previously acquired knowledge~\cite{ke2023cpt_lm, chen2025cpt_comp, gupta2023cpt_warmup, yildiz2024investigating_cpt}. By reducing optimization conflicts in subsequent alignment stages, agentic CPT significantly improves overall agentic performance. The training data in this stage typically consists of large-scale synthetic corpora, and the optimization objective is based on Maximum Likelihood Estimation:
\begin{equation}
\mathcal{L}_{\mathrm{MLE}}(\theta) 
= - \sum_{t=1}^{T} \log p_\theta(x_t \mid x_{<t}),
\end{equation}
\noindent where $x_t$ denotes the target token at time step $t$, $x_{<t}$ represents the preceding sequence of tokens, and $p_\theta$ is the conditional probability distribution over the next token parameterized by $\theta$.

\subsection{Agentic Supervised Fine-tuning}  
Agentic Supervised Fine-tuning (Agentic SFT) is typically introduced as an initialization stage before reinforcement learning~\cite{wu2025masksearch, wu2025webwalker, li2025websailor-v2, li2025websailor, tao2025webshaper,yao2023react}, providing a strong policy prior by leveraging high-quality datasets. These datasets contain detailed agentic trajectories, often synthesized through reverse engineering~\cite{wu2025webwalker}, graph-based synthesis~\cite{li2025websailor, li2025websailor-v2}, and formalized task modeling~\cite{tao2025webshaper}.
The goal of agentic SFT is to align the model with action execution patterns, specifying what actions to perform and how to carry them out effectively. The optimization objective of agentic SFT remains Maximum Likelihood Estimation, consistent with Agentic CPT, though the two stages differ in both their data characteristics and training purposes.

\subsection{Agentic Reinforcement Learning}
Agentic Reinforcement Learning (Agentic RL) is a post-training paradigm that leverages exploration and reward-based feedback to refine agentic capabilities.
Its core objective is to maximize the expected cumulative reward by iteratively refining planning processes and optimizing decision policies.
We next introduce two classic RL algorithms widely used in Agentic RL, i.e., PPO~\cite{ppo} and GRPO~\cite{shao2024deepseekmath}.

\textbf{Proximal Policy Optimization (PPO).}  
PPO~\cite{ppo} is an actor–critic RL algorithm for aligning models with desired behaviors, which refines the policy through iterative updates that promote exploration while constraining excessive deviation from the previous policy.  
This balance is achieved via a clipped objective, which stabilizes optimization and mitigates the risk of performance collapse.
Formally, given a policy $\pi_\theta$, a previous policy $\pi_{\theta_{\text{old}}}$, and an advantage estimator $A_t$, the PPO objective is defined as:  
\begin{align}
&\mathcal{J}_{PPO}(\theta) 
= \mathbb{E}_{(I,T)\sim p_{\mathcal{D}},\, o\sim\pi_{\theta_\text{old}}} \nonumber\\
&\quad \frac{1}{|o|} \sum_{t=1}^{|o|} 
\min\Bigg(
\frac{\pi_\theta(o_t \mid I,T)}{\pi_{\theta_{\text{old}}}(o_t \mid I,T)} A_t, \;
\mathrm{clip}\Bigg(
\frac{\pi_\theta(o_t \mid I,T)}{\pi_{\theta_{\text{old}}}(o_t \mid I,T)},\, \nonumber\\
& ~~~~~ 1-\epsilon,\,1+\epsilon
\Bigg) A_t
\Bigg).
\end{align}
where $\epsilon$ is the clipping parameter that bounds policy updates, and $A_t$ is the advantage, often estimated using Generalized Advantage Estimation (GAE). To further encourage linguistic coherence and mitigate reward hacking, a KL divergence penalty relative to a reference model $\pi_{\text{ref}}$ is commonly added to the reward:  
\begin{equation}
r_t = r_\varphi(q, o_{\leq t}) 
- \beta \log \frac{\pi_\theta(o_t \mid q, o_{<t})}{\pi_{\text{ref}}(o_t \mid q, o_{<t})},
\end{equation}
where $r_\varphi$ denotes the reward model and $\beta$ controls the regularization strength.

\textbf{Group Relative Policy Optimization (GRPO).} 
GRPO is a simplified variant of PPO that removes the need for a separate value function.
It estimates the baseline directly from rollouts, reducing the cost of training a value model while maintaining stable policy updates.
For each question $q$, GRPO samples a group of responses $\{o_1, o_2, \ldots, o_G\}$ from the old policy $\pi_{\theta_{\text{old}}}$, with rewards $\{R_1, R_2, \ldots, R_G\}$ assigned by rules or models.
The rewards are then normalized by subtracting the group mean and dividing by the standard deviation to obtain the relative advantage for each response:
\begin{align}
\hat{A}_{i} = \frac{R_i - \text{mean}\left(\{R_j\}_{j=1}^{G}\right)}{\text{std}\left(\{R_j\}_{j=1}^{G}\right)}.
\end{align}
Based on these normalized advantages, the training objective is defined as:  
\begin{align}
&\mathcal{J}_{\mathrm{GRPO}}(\theta)
= \mathbb{E}_{(I,T)\sim p_{\mathcal{D}},\, o\sim\pi_{\theta_\text{old}}(\cdot|I,T)} \nonumber\\
&\quad \Biggl[ \frac{1}{n}\sum_{i=1}^{n} 
\min\!\Biggl(
\frac{\pi_{\theta}(o_i \mid I,T)}{\pi_{\theta_{\mathrm{old}}}(o_i \mid I,T)}\hat{A}_i,\;
\mathrm{clip}\!\Bigl(
\frac{\pi_{\theta}(o_i \mid I,T)}{\pi_{\theta_{\mathrm{old}}}(o_i \mid I,T)},\, \nonumber\\
& ~~~~1-\epsilon,\,1+\epsilon
\Bigr)\hat{A}_i
\Biggr) 
- \beta D_{\mathrm{KL}}\!\left(\pi_{\theta}\,\|\, \pi_{\mathrm{ref}}\right)
\Biggr]\textrm{,}
\end{align}
where $\hat{A}_i$ is the normalized advantage of candidate $o_i$, $\pi_{\theta}$ is the current policy, $\pi_{\theta_{\text{old}}}$ denotes the previous policy, $\pi_{\mathrm{ref}}$ is a reference policy for KL regularization, and $\epsilon$ and $\beta$ control clipping and regularization strength, respectively.


\subsection{Agentic Evaluation}
Agentic MLLMs generate long-horizon action trajectories when solving complex problems.
Accordingly, the evaluation can be categorized into two complementary dimensions: process evaluation and outcome evaluation.

\textbf{Process Evaluation.} This dimension focuses on whether the agentic MLLM can generate accurate intermediate processes, such as precise reasoning steps~\cite{zhang2024mathverse, jiang2025mme_cot, yao2025mmreason} or appropriate tool invocations~\cite{huang2024tool_bench, huang2023metatool, li2025mm-BrowseComp}. It assesses the logical consistency of reasoning paths and the appropriateness of tool usage, thereby reflecting the transparency, reliability, and robustness of the intermediate process.

\textbf{Outcome Evaluation.} This dimension measures the ability of agentic MLLMs to produce accurate and helpful solutions across diverse downstream tasks~\cite{mathvista, mathvision, liu2025screens, rawles2024androidworld}. It reflects their generalization ability, and problem-solving competence as agentic systems.

Together, these two dimensions provide a comprehensive framework for evaluating agentic MLLMs, capturing both the quality of their intermediate processes and the effectiveness of their final outcomes.

\section{Agentic MLLM}

\label{Sec: 4 agentic mllm}
In this section, we categorize agentic MLLMs into three core components: \textbf{internal intelligence} (Section~\ref{Sec: 4.1 internal self-planning}), \textbf{external tool invocation} (Section~\ref{Sec: 4.2 external tool}), and \textbf{environmental interaction} (Section~\ref{Sec: 4.3 environment interaction}), as illustrated in Figure~\ref{fig:3 overview of survey}.
First, internal intelligence constitutes the cognitive core of agentic MLLMs, comprising long-chain reasoning (Section~\ref{sec: 4.1.1 agentic reasoning}), reflection (Section~\ref{sec: 4.1.2 agentic reflection}), and memory (Section~\ref{sec: 4.1.3 agentic memory})
Internal intelligence enables the model to construct coherent chains of reasoning and strategic plans, orchestrating subsequent actions to accomplish tasks step by step.
Second, under the coordination of internal intelligence, agentic MLLMs can proactively invoke various external tools to acquire required information (Section~\ref{sec: 4.2.1 agentic search}), execute code for complex computations (Section~\ref{sec: 4.2.2 agentic code}), and process visual representations to strengthen reasoning (Section~\ref{sec: 4.2.3 agentic data process}). This human-like tool use substantially extends their problem-solving capabilities beyond intrinsic knowledge.
Finally, with deliberate planning and tool use, agentic MLLMs interact with both virtual (Section~\ref{sec: 4.3.1 agentic Virtual Embodiment}) and physical environments (Section~\ref{sec: 4.3.1 agentic Physical Embodiment}). Through such interactions, agentic MLLMs can perceive external environments and receive feedback, enabling dynamic adaptation in real-world deployments.

\subsection{Agentic Internal Intelligence}
\label{Sec: 4.1 internal self-planning}
\begin{table*}[!t]
    \setlength\tabcolsep{6pt}
    \centering
    \caption{Summary of agentic internal intelligence, grouped into three categories: Reasoning, Reflection and Memory. 
    }
    \resizebox{0.98\linewidth}{!}{
    \begin{tabular}{lccp{15cm}}
    \toprule[1pt]
    \rowcolor{red!5}
        \textbf{Reasoning} & \textbf{Fine-tuning} &  \textbf{Reward Modeling} & \textbf{Contribution} \\
    \midrule


        
    


        Vision-R1~\cite{huang2025visionr1}~\href{https://github.com/Osilly/Vision-R1}{[code]} & SFT+RL & Outcome + Rule &    Introduce progressive thinking suppression training within GRPO to progressively optimize the model. \\ 

        MM-Eureka~\cite{meng2025mmekura} ~\href{https://github.com/ModalMinds/MM-EUREKA}{[code]} & SFT+RL & Outcome + Rule & Introduce high-quality MM-K12 with online filtering and a two-stage training strategy to improve stability. \\ 

        Skywork R1V2~\cite{wang2025skyworkr1v2}~\href{https://huggingface.co/Skywork/Skywork-R1V2-38B}{[code]}      
        & SFT+RL & Outcome + Rule
        & Propose selective sample buffer to tackle GRPO’s vanishing advantages by prioritizing high-value samples. \\

        Video-R1~\cite{feng2025videor1}~\href{https://github.com/tulerfeng/Video-R1}{[code]} & SFT+RL & Outcome + Rule & Construct 165k cold start and 260k RL dataset; Propose T-GRPO to explicitly encourage temporal reasoning in videos. \\

        LongVILA-R1~\cite{chen2024longvila}~\href{https://github.com/NVlabs/Long-RL}{[code]} & SFT+RL & Outcome + Rule & Introduce 104K long-video QA pairs with reasoning annotations and a two-stage pipeline of CoT-SFT and RL.  \\

        ThinkLiteVL~\cite{wang2025thinklitevl} ~\href{https://github.com/si0wang/ThinkLite-VL}{[code]} & RL & Outcome + Rule & Repurpose MCTS to identify challenging yet solvable examples that enhance RL effectiveness in low-data regimes. \\

        R1-ShareVL~\cite{yao2025r1sharevl}~\href{https://github.com/HJYao00/R1-ShareVL}{[code]} & RL & Outcome + Rule & Expand the question space and shares reasoning trajectories and rewards across variants to mitigate sparse rewards. \\
        EchoInk-R1~\cite{xing2025Echoink-r1} & RL & Outcome + Rule & A GRPO framework for audio–image QA, showing reflection by revisiting and refining responses under ambiguity.\\
        Infi-MMR~\cite{liu2025infi-mmr}~\href{https://huggingface.co/InfiX-ai/Infi-MMR-3B}{[code]} & RL & Outcome + Rule & A curriculum learning activating reasoning with text, adapting with captions and enhancing with caption-free data. \\
        NoisyRollout~\cite{liu2025noisyrollout}~\href{https://github.com/John-AI-Lab/NoisyRollout}{[code]}  & RL & Outcome + Rule & Augment RL by mixing clean and distorted trajectories with noise annealing to improve exploration and robustness. \\
        VL-Cogito~\cite{yuan2025vlcogito}~\href{https://github.com/alibaba-damo-academy/VL-Cogito}{[code]} & RL & Outcome + Rule & Curriculum RL with difficulty soft weighting and dynamic length rewards to balance efficiency and correctness. \\

        WeThink~\cite{yang2025wethink}~\href{https://github.com/yangjie-cv/WeThink}{[code]} & RL & Outcome + Model & Present a hybrid reward combining rule-based verification and model-based assessment to optimize RL across tasks.\\

        R1-VL~\cite{zhang2025r1vl} ~\href{https://github.com/jingyi0000/R1-VL}{[code]} & SFT+RL & Process + Rule & Introduce two rewards to help models cover key intermediate steps while maintaining structural and logical consistency. \\ 

        SophiaVL-R1~\cite{fan2025sophiavl}~\href{https://github.com/kxfan2002/SophiaVL-R1}{[code]} & SFT+RL & Process + Model & Introduce process-level rewards by a trained reward model and using Trust-GRPO to weight their reliability. \\

        Perception-R1~\cite{perception_r1}~\href{https://github.com/tongxiao2002/Perception-R1}{[code]} & RL & Process + Model & Propose a visual perception reward, judged by an LLM for annotation–response consistency. \\

        GRPO-CARE~\cite{chen2025grpocare}~\href{https://github.com/TencentARC/GRPO-CARE}{[code]} & RL & Process + Model & Consistency-aware learning with correctness rewards and an adaptive consistency bonus for coherent reasoning. \\

    \midrule
        \rowcolor{red!5}
        \textbf{Reflection} & \textbf{Trigger Type} &  \textbf{Reflection Granularity} & \textbf{Contribution} \\
        \midrule
        VLAA-Thinker~\cite{vlaathinker}~\href{https://ucsc-vlaa.github.io/VLAA-Thinking/}{[code]} & Implicit & Step level & Demonstrate that GRPO training induces reflection, evidenced by the frequency of four “aha” expressions. \\
        MM-Eureka~\cite{meng2025mmekura} ~\href{https://github.com/ModalMinds/MM-EUREKA}{[code]} & Implicit & Step level & RL optimization induces reflection in MLLMs without explicit incentives. \\
        FRANK~\cite{frank}~\href{http://iip.whu.edu.cn/frank/index.html}{[code]} & Implicit & Step level & Propose hierarchical weight merging of a MLLM and a reasoning-specialized LLM, revealing emergent reflection. \\
        Mulberry~\cite{yao2024mulberryempoweringmllmo1like}~\href{https://github.com/HJYao00/Mulberry/tree/main}{[code]} & Explicit & Step level & Leverage CoMCTS to build reflective reasoning paths by incorporating negative sibling nodes into trajectories. \\
        Vision-R1~\cite{huang2025visionr1}~\href{https://github.com/Osilly/Vision-R1}{[code]} & Explicit & Step level & Introduce a cold-start dataset, vision-r1-cold, featuring a higher frequency of reflective markers. \\
        VL-Rethinker~\cite{wang2025vlrethinker}~\href{https://github.com/TIGER-AI-Lab/VL-Rethinker}{[code]} & Explicit & Step level & Explicit “rethinking triggers” during rollouts, guiding VLMs toward strategic reflection. \\
        Gthinker~\cite{zhan2025gthinker}~\href{https://github.com/jefferyZhan/GThinker}{[code]} & Explicit & Step level & Propose a reasoning pattern that grounds in visual cues and iteratively reinterprets them to resolve inconsistencies. \\
        R$^3$V~\cite{r3v_reflection}~\href{https://github.com/njucckevin/MM-Self-Improve}{[code]} & Explicit & Response level & Iteratively generate positive/negative solutions, apply self-reflection to refine flaws, and select superior reasoning paths. \\
        SRPO~\cite{wan2025srpo}~\href{https://srpo.pages.dev/}{[code]} & Explicit & Response level & Proposes a two-stage framework to enhance reasoning with reflection-focused data and a reflection-aware GRPO reward. \\
        Look-Back~\cite{yang2025Look-Back}~\href{https://github.com/PKU-YuanGroup/Look-Back}{[code]} & Explicit & Response level & Introduce an implicit method enabling MLLMs to self-reflect by re-focusing on visual inputs during reasoning. \\
        LongVIL~\cite{chen2025longvil}~\href{https://longvil-agent.github.io/}{[code]} & Explicit & Response Level & An agent with plan and code reflection to refine actions and code, ensuring temporal-spatial coherence and correctness. \\

    \midrule
        \rowcolor{red!5}
        \textbf{Memory} & \textbf{Memory Type} &  \textbf{Mechanism} & \textbf{Contribution} \\
        \midrule
        BLIP-2~\cite{li2023blip2}~\href{https://github.com/salesforce/LAVIS/tree/main/projects/blip2}{[code]} & Contextual & Token Compression & Leverage a two-stage pretrained Querying Transformer to bridge the modality gap and compress visual tokens. \\
        Dense Connector~\cite{yao2024denseconnector}~\href{https://github.com/HJYao00/DenseConnector}{[code]} & Contextual & Token Compression & Use a parameter-free connector layer to compress visual tokens, accelerating inference while preserving performance. \\
        Qwen2.5-VL~\cite{bai2025qwen2.5vl}~\href{https://github.com/QwenLM/Qwen3-VL}{[Code]} & Contextual & Token Compression & An MLP compresses adjacent visual patch features into the text embedding space for efficient vision-language fusion. \\
        LongRoPE~\cite{ding2024longrope}~\href{https://github.com/microsoft/LongRoPE}{[code]} & Contextual &  Window Extension & Extend window to 2048K by non-uniform interpolation search, progressive extension training, LongRoPE readjustment. \\
        LongLM~\cite{jin2024longlm}~\href{https://github.com/datamllab/LongLM}{[code]} & Contextual &  Window Extension &  Extend context window by constructing bi-level attention information: the grouped attention and the neighbor attention. \\
        LongVA~\cite{zhang2024longva}~\href{https://github.com/EvolvingLMMs-Lab/LongVA}{[code]} & Contextual &  Window Extension &  Extrapolate LLM’s context length, enabling MLLMs to comprehend orders of magnitude more visual tokens.  \\
        LongVILA~\cite{chen2024longvila}~\href{https://github.com/NVlabs/VILA/tree/main/longvila}{[code]} & Contextual &  Window Extension &  Upgrade VLMs to support long context understanding by long context extension and long video SFT. \\
        S$^2$CAN~\cite{s2can} & External & Heuristic-driven & A memory-augmented framework enhancing surgical context understanding with direct and indirect memories.\\
        MA-LMM~\cite{he2024malmm}~\href{https://github.com/boheumd/MA-LMM}{[code]} & External & Heuristic-driven & Design distinct visual and query memory banks to separately manage information from different modalities. \\
        MovieChat~\cite{song2024moviechat}~\href{https://github.com/rese1f/MovieChat}{[code]} & External & Heuristic-driven & Combine a sliding-window short-term memory with a compact long-term memory to consolidate video tokens. \\
        MemoryBank~\cite{zhong2023memorybank} & External & Heuristic-driven & Evolve memories, adapt to users, and use an Ebbinghaus-inspired mechanism to forget or reinforce information. \\
        MemTool~\cite{lumer2025memtool} & External & Heuristic-driven & 
        Short-term memory for tool/context control in multi-turn conversations with autonomous, workflow and hybrid modes. \\
        A-Mem~\cite{xu2025a-mem}~\href{https://github.com/WujiangXu/A-mem-sys}{[code]} & External & Reasoning-driven &
        A Zettelkasten-inspired memory system building evolving knowledge networks via dynamic indexing and linkin.\\
        MEM1~\cite{zhou2025mem1}~\href{https://github.com/MIT-MI/MEM1}{[code]} & External & Reasoning-driven & An RL framework maintaining constant memory in multi-turn tasks via compact updates and redundancy reduction. \\
        Memory-R1~\cite{yan2025memoryr1}  & External & Reasoning-driven & 
        An RL-based memory manager and answer agent for adaptive external memory management beyond static heuristics.\\

        RMM~\cite{tan2025prospect_rmm} & External & Reasoning-driven &  Long-term dialogue with Prospective Reflection for memory and Retrospective Reflection for RL-based refinement. \\ 
        M+~\cite{wang2025m+}~\href{https://github.com/wangyu-ustc/MemoryLLM}{[code]} & External & Reasoning-driven & A memory-augmented model with long-term memory and a co-trained retriever for dynamic retrieval during generation. \\        
        
    \bottomrule[1pt]
    \end{tabular}
    }
    \label{tab: 1 agentic internal intellligence}
\end{table*}
Agentic internal intelligence denotes the capacity of a model to deliberately organize and coordinate actions in pursuit of a goal, forming the cornerstone of effective task execution.
For MLLMs, achieving such internal intelligence relies on the integration of three complementary abilities: \textbf{reasoning}, \textbf{reflection}, and \textbf{memory}.
These abilities collectively enable the model to coherently construct, validate, and refine its decision-making process, maintaining consistency across extended agentic trajectories.
To this end, this section reviews recent approaches to advancing internal intelligence in MLLMs along these three dimensions. A summary of internal intelligence method is provided in Table~\ref{tab: 1 agentic internal intellligence}.

\subsubsection{Agentic Reasoning}
\label{sec: 4.1.1 agentic reasoning}
Agentic reasoning in MLLMs refers to the deliberate generation of intermediate reasoning steps prior to producing a final answer, a process that substantially enhances their capacity to tackle complex problems~\cite{jaech2024openai-o1, guo2025deepseek-r1}.
Current efforts to strengthen reasoning capabilities can be broadly categorized into three learning paradigms: \textbf{prompt-based reasoning}, \textbf{SFT-based Reasoning}, and \textbf{RL-based reasoning}. Each paradigm is introduced in the following.

\textbf{Prompt-based Reasoning.}
Prompt-based approaches guide MLLMs to generate explicit intermediate reasoning steps by incorporating instructions such as ``Let us solve the problem step by step''~\cite{cot, llm_zero-shot_reason}.
This strategy encourages models to articulate multi-step reasoning trajectories before arriving at a final answer and has been shown to improve performance on complex tasks across diverse domains.

Building on this foundation, subsequent work has extended prompt-based CoT reasoning along both depth and breadth.  
Best-of-N (BoN) methods independently generate multiple reasoning paths and then select the best one using either a reward model~\cite{wang2025visualprm, du2025mmprm, chen2025rm-r1} or heuristic scoring functions~\cite{bon_self-certainty, bon_speculative}. Representative studies such as VisualPRM~\cite{wang2025visualprm}, MM-PRM~\cite{du2025mmprm}, and RM-R1~\cite{chen2025rm-r1} train specialized reward models to better evaluate and select reasoning trajectories.  
Tree search methods~\cite{tot, fot} further extend CoT by expanding reasoning paths into tree structures, allowing structured exploration beyond linear chains. VisuoThink~\cite{wang2025visuothink}, for instance, enables multimodal slow thinking through progressive visual–textual reasoning, while incorporating test-time scaling via look-ahead tree search.
Furthermore, Monte Carlo Tree Search (MCTS)~\cite{MCTS} introduces a principled balance between exploration and exploitation by progressively expanding promising branches through stochastic rollouts and statistical evaluation. Building on this, AStar~\cite{astar} applies MCTS-derived thought cards to achieve more structured reasoning at test time.  

Despite their empirical successes, prompt-based methods remain fundamentally constrained by the fixed knowledge encoded in model parameters and the limited search space available at inference. These limitations restrict their scalability and robustness when applied to more challenging, open-ended tasks.

\textbf{\textit{SFT-based Reasoning.}} 
Supervised Fine-Tuning (SFT) on long-chain reasoning datasets compels MLLMs to learn reasoning abilities by minimizing the MLE loss over annotated reasoning traces.
The central challenge lies in constructing high-quality reasoning datasets. We broadly categorize these approaches by their synthesis methodologies and introduce them below.

Direct distillation is a simple yet widely used method that generates reasoning paths directly from stronger teacher models, exemplified by LLaVA-Reasoner~\cite{llava-reasoner}, MAmmoTH-VL~\cite{Mammoth-vl}, and MAVIS~\cite{zhang2024mavis}.
Structured distillation decomposes the reasoning process into predefined modules to reduce question complexity, and then instructs the powerful model to generate each component in sequence; for example, LLaVA-CoT~\cite{xu2025llavacotletvisionlanguage} partitions reasoning into four stages: summary, caption, reasoning, and conclusion.  
Tree distillation treats each reasoning step as a node in a tree, forcing the model to generate and explore multiple branches before pruning less promising ones to obtain higher-quality reasoning traces. Mulberry~\cite{yao2024mulberryempoweringmllmo1like} introduces collective learning into MCTS to more effectively search reasoning and reflection trajectories.

Recent works~\cite{chen2024m3cot, zhang2023multimodal, jiang2025corvid, luan2024textcot, shao2024visualcot, dong2025insightv, ccot, thawakar2025llamavo1rethinkingstepbystepvisual, wang2025r1-compress} utilize these reasoning datasets to fine-tune MLLMs, advancing the development of reasoning MLLMs.
Nevertheless, the reliance on high-quality CoT reasoning paths and the constrained learning mechanism of SFT, which often ties MLLMs to fixed reasoning patterns, remains a major challenge for achieving generalizable reasoning.

\textbf{RL-based Reasoning.}
A major breakthrough in MLLM reasoning was marked by efforts such as OpenAI o1~\cite{jaech2024openai-o1} and DeepSeek R1~\cite{guo2025deepseek-r1}, which applied large-scale reinforcement learning and achieved transformative gains.  
By leveraging exploration and feedback signals, RL optimizes reasoning trajectories, allowing MLLMs to reason in a more flexible, adaptive, and dynamic manner.
For long-chain reasoning, reward modeling in RL is typically divided into two paradigms: outcome-based rewards, which evaluate only the final answers, and process-based rewards, which additionally assess intermediate reasoning steps.  
Both can be assigned rewards through either rule-based heuristics or specialized reward models. In the following, we review representative methods according to their reward formulations.

\begin{itemize}
    \item \textbf{Outcome reward modeling} assigns rewards based solely on final answer correctness, ignoring the intermediate reasoning process. It is simple to implement and has attracted widespread attention, particularly following the success of DeepSeek-R1~\cite{guo2025deepseek-r1}, which employed rule-based reward computation to mitigate reward-model hacking~\cite{Reward-Hacking, wang2025beyondrewardhacking} and to lower the need for additional training resources.
    Subsequent work~\cite{deng2025openvlthinkerearlyexplorationcomplex, shen2025vlm-r1, liu2025segzero, peng2025lmm-r1, zhan2025visionr1, yin2025tinyr1vlightweightmultimodalunified} extended outcome-based RL to the multimodal domain.
    Early work Vision-R1~\cite{huang2025visionr1} introduces progressive thinking suppression training into GRPO, mitigating token explosion and improving training stability.
    Recent works enhance multimodal reasoning capabilities through techniques such as high-quality data selection~\cite{wang2025thinklitevl, yuan2025vlcogito} and data augmentation~\cite{liu2025noisyrollout, yao2025r1sharevl}, addressing advantage vanishing~\cite{yao2025r1sharevl, wang2025skyworkr1v2, wang2025vlrethinker, huang2025mapo} and curriculum learning~\cite{yuan2025vlcogito, liu2025infi-mmr, Curr-ReFT}.
    Beyond rule-based judgment, WeThink~\cite{yang2025wethink} has also introduced reward models to verify the correctness of final answers.
    
    \item \textbf{Process reward modeling} extends outcome-based rewards by incorporating supervision at the intermediate step level in addition to outcome-level evaluation, guiding intermediate reasoning steps to improve the quality and robustness of the reasoning process.
    R1-VL~\cite{zhang2025r1vl} introduces rule-based process rewards by matching extracted keywords from reasoning steps to predefined rules, enabling finer-grained control and alleviating advantage vanishing.
    Other works, such as Perception-R1~\cite{perception_r1}, SophiaVL-R1~\cite{fan2025sophiavl}, and GRPO-CARE~\cite{chen2025grpocare}, employ specialized process reward models to score intermediate steps, improving reasoning reliability, coherence, and consistency.

\end{itemize}

In summary, outcome-based rewards are simple to implement and efficient to scale, but they overlook the reasoning process. In contrast, process-based rewards provide finer-grained supervision that improves reliability and coherence, though they demand more complex design, incur higher computational costs, and remain vulnerable to reward model hacking.

\subsubsection{Agentic Reflection}
\label{sec: 4.1.2 agentic reflection}
MLLMs are inherently constrained by the autoregressive paradigm, where errors are irreversible and tend to accumulate over time.  
Drawing inspiration from its central role in human cognition, reflection has been introduced into LLMs as a mechanism to overcome this limitation.  
Recent studies~\cite{shinn2023reflexion, yao2024mulberryempoweringmllmo1like, r3v_reflection} demonstrate that reflective strategies enable models to verify and refine their responses, thereby enhancing robustness, mitigating hallucinations, and supporting more effective agentic internal intelligence.
The approaches for inducing reflection can be categorized into explicit and implicit methods.

\textbf{Implicitly Induced Reflection.}
Studies such as DeepSeek-R1~\cite{guo2025deepseek-r1} have observed that models can exhibit emergent reflective behaviors after reinforcement learning.
These reflective behaviors are not explicitly induced, but rather emerged organically through interaction with the reinforcement learning exploration.
Similar emergent reflections have also been reported in MLLMs, as shown by MM-Eureka~\cite{meng2025mmekura} and VLAA-Thinker~\cite{vlaathinker}.

\textbf{Explicitly Induced Reflection.}  
Subsequent research~\cite{yao2024mulberryempoweringmllmo1like,zhan2025gthinker,wan2025srpo} introduces mechanisms that explicitly induce reflective behaviors in MLLMs.
These methods can be broadly divided into two categories: \textbf{response-level} reflection, which is applied after the model generates a complete response, and \textbf{step-level} reflection, which is introduced during intermediate reasoning steps.

\begin{itemize}
    \item \textbf{Response-level reflection.}
    In this setting, reflection is triggered only after the model generates a complete response, which can be formalized as:  
    \begin{align}
    response &= r^{-} \; + \; \rho \; + \; r^{+},
    \end{align}
    \noindent where $r^{-}$ denotes the initial flawed response, $r^{+}$ represents the refined response, and $\rho$ is the reflection prompt linking the two.  
    Representative methods include R$^{3}$V~\cite{r3v_reflection}, which fosters reflective capability by iteratively generating positive and negative solutions, applying self-reflection losses to refine flawed rationales, and selecting superior reasoning paths.  
    SRPO~\cite{wan2025srpo} introduced a two-stage RL framework that leverages reflection-enhanced data and reflection-aware GRPO rewards to incentivize reflective behaviors.

    \item \textbf{Step-level reflection.}
    In this setting, reflection is interleaved between intermediate reasoning steps so that each draft step is critiqued and revised before proceeding, formalized as:
    \begin{align}
    response &= s_{1} \; + \; s_{2}^{-} \; + \; \rho \; + \; s_{2}^{+} \; + \cdots + \; s_{n},
    \end{align}
    \noindent where $s_t^{-}$ denotes the $t$-th initial flawed reasoning step, $s_t^{+}$ represents the revised step after reflection, and $\rho$ indicates the reflection prompt inserted between consecutive steps.
    Mulberry~\cite{yao2024mulberryempoweringmllmo1like} exemplifies this paradigm by employing collective MCTS to construct reflective reasoning paths, explicitly incorporating negative sibling nodes to incentivize reflection.
    VL-Rethinker~\cite{wang2025vlrethinker} advances this direction by designing explicit rethinking triggers during rollouts, guiding MLLMs toward more strategic reflection.

\end{itemize}

\subsubsection{Agentic Memory}
\label{sec: 4.1.3 agentic memory}
Memory plays a pivotal role in advancing MLLMs beyond the limitations of the fixed and limited context window.
By retaining and leveraging past information, it enables models to maintain continuity across sessions, and support more coherent internal intelligence over long-horizon interactions.
In this section, we divide memory into \textbf{contextual} and \textbf{external} memory systems for detailed discussion. The summary of agentic memory research is provided in Figure~\ref{tab: 1 agentic internal intellligence}

\textbf{Contextual Memory.}
Contextual memory refers to directly concatenating past information into the current context window, providing a simple yet effective way to leverage history for response generation.  
However, the fixed context length imposes strict limits, motivating two primary strategies: \textbf{token compression} and \textbf{window extension}.  

\begin{itemize}

\item  \textbf{Token compression.} This strategy reduces the number of tokens by condensing input representations, thereby indirectly increasing the effective capacity of the context window.  
Parametric methods typically employ a Query Transformer (Q-Former) to downsample high-dimensional features into a smaller set of informative learnable tokens.  
Representative works include Flamingo~\cite{alayrac2022flamingo}, BLIP-2~\cite{li2023blip2}, and Video-LLaMA~\cite{zhang2023video-llama}, which use Q-Former as a vision–language bridge to efficiently compress visual inputs.  
In contrast, non-parametric approaches rely on traditional pooling operations (e.g., average pooling or max pooling)~\cite{lecun1989handwritten_pooling}.  
Such methods have been explored in PLLaVA~\cite{xu2024pllava} and Dense Connector~\cite{yao2024denseconnector}, where pooling is applied to compress multimodal inputs without introducing additional learnable parameters.

\item \textbf{Window Extension.} Unlike token compression, which enlarges context capacity indirectly, another line of work focuses on directly extending the context window.
LongRoPE~\cite{ding2024longrope} expands the original context length from 128k to 2048k tokens through a progressive extension strategy.  
In the multimodal domain, LongVILA~\cite{chen2024longvila} and LongVA~\cite{zhang2024longva} extend the context window to handle inputs exceeding 2,000 video frames, supporting long-horizon temporal reasoning.

\end{itemize}

\textbf{External Memory Systems.}
Some studies~\cite{zhong2023memorybank, memgpt} extend memory beyond the internal context window by incorporating external modules for storing and retrieving information.  
Based on their mechanisms, these approaches can be broadly divided into heuristic-driven and reasoning-driven memory systems. 

\begin{itemize}

\item \textbf{Heuristic-driven memory.}  
Early external memory systems relied on static, rule-based pipelines with predefined strategies for storing, updating, and retrieving information.  
For example, MemoryBank~\cite{zhong2023memorybank} and MemGPT~\cite{memgpt} use specialized prompts to manage textual memory, while MovieChat~\cite{song2024moviechat} and MovieChat+~\cite{song2024moviechat+} introduce both short-term and long-term modules to process videos exceeding 10K frames.
Similarly, MA-LMM~\cite{he2024malmm} maintains separate memory banks for visual and query information.  
Although effective in constrained domains, these systems depend on fixed heuristics, which limit adaptability in dynamic and open-ended environments.  

\begin{table*}[!ht]
    \setlength\tabcolsep{6pt}
    \centering
    \caption{Summary of agentic external tool invocation, grouped into three categories: Search, Coding and Visual Processing. 
    }
    \resizebox{0.98\linewidth}{!}{
    \begin{tabular}{lccp{16cm}}
    \toprule[1pt]
    \rowcolor{red!5}
        \textbf{Agentic Search} & \textbf{Fine-tuning} &  \textbf{Search Modalities} & \textbf{Contribution} \\
    \midrule
        MMSearch~\cite{jiang2024mmsearch}~\href{https://mmsearch.github.io/}{[code]} & Prompt-based & T, I & Introduce multimodal AI search engine pipeline that equips MLLMs with multimodal search capabilities. \\
        Search-R1~\cite{jin2025searchr1}~\href{https://github.com/PeterGriffinJin/Search-R1}{[code]} & RL & T & RL framework to autonomously generate multi-turn search queries stabilized by retrieved token masking and outcome reward. \\
        VRAG-RL~\cite{wang2025vrag-rl}~\href{https://github.com/Alibaba-NLP/VRAG}{[code]} & SFT+RL & I & 

        Visual action space with cropping–scaling for info gathering and a reward uniting query rewriting and retrieval performance.
        \\
        Visual-ARFT~\cite{liu2025visualarft}~\href{https://github.com/Liuziyu77/Visual-RFT/tree/main/Visual-ARFT}{[code]} & RL & T, I & Enable MLLMs to flexibly reason by browsing websites for real-time information and coding adaptive image manipulations. \\
        MM-Search-R1~\cite{wu2025mmsearchr1}~\href{https://github.com/EvolvingLMMs-Lab/multimodal-search-r1}{[code]} & SFT+RL & T, I & 
        Learn when and how to perform image–text search by SFT and RL, guided by outcome-based rewards with a search penalty.
        \\
        M2IO-R1~\cite{xiao2025m2io-r1}& SFT+RL & T, I & A MRAG framework supporting multimodal I/O with controllable, semantically aligned image selection and placement. \\
        Patho-AgenticRAG~\cite{zhang2025Patho-AgenticRAG}~\href{https://github.com/Wenchuan-Zhang/Patho-AgenticRAG}{[code]} & SFT+RL & T, I & Page-level embedding database for text–image retrieval with reasoning, decomposition, and multi-turn search in diagnostics. \\
        WebWatcher~\cite{geng2025webwatcher}~\href{https://github.com/Alibaba-NLP/WebAgent}{[code]} & SFT+RL & T, I & Leverage synthetic multimodal trajectories for efficient cold-start, enabling tool use and improved generalization via RL. \\

    \midrule
        \rowcolor{red!5}
        \textbf{Agentic Coding} & \textbf{Fine-tuning} &  \textbf{Application} & \textbf{Contribution} \\
        \midrule

        Posterior-GRPO~\cite{fan2025posterior} & RL & Programming & Introduce Posterior-GRPO and tailored reward models to guide intermediate reasoning for more accurate code generation. \\
        
        R1-Code-Interpreter~\cite{chen2025r1}~\href{https://github.com/yongchao98/R1-Code-Interpreter}{[code]} & SFT+RL & Programming & Achieve successful multi-turn interleaved textual reasoning and code generation across multiple tasks.\\
        ToRA~\cite{gou2023tora}~\href{https://github.com/microsoft/ToRA}{[code]} & SFT & Mathematics & A pioneering line of work that integrates external coding tools into the textual reasoning process.\\
        MathCoder~\cite{wang2023mathcoder}~\href{https://github.com/mathllm/MathCoder}{[code]} & SFT & Mathematics & Present MathCodeInstruct, a high-quality SFT dataset, and MathCoder, a family of models for mathematical reasoning. \\
        
        rStar-Math~\cite{guan2025rstar}~\href{https://github.com/microsoft/rStar}{[code]} & SFT & Mathematics & Propose a self-evolution framework that integrates an MCTS-based data synthesis method with a process preference model.\\

        ToRL~\cite{li2025torl}~\href{https://github.com/GAIR-NLP/ToRL}{[code]} & RL & Mathematics & Achieve tool-integrated reasoning on challenging mathematical problems through reinforcement learning. \\
        Retool~\cite{feng2025retool}~\href{https://github.com/ReTool-RL/ReTool}{[code]} & SFT+RL & Mathematics & Constructs an outcome-driven RL framework for multi-turn tool invocation and long-form reasoning. \\
        
        OTC~\cite{wang2025otc} & RL & Mathematics & Incentivize models to solve tasks correctly using minimal tool interactions via Optimal Tool Call-controlled Policy Optimization.\\
        
        CoRT~\cite{li2025cort}~\href{https://github.com/ChengpengLi1003/CoRT}{[code]} & SFT+RL & Mathematics &Propose a hint-engineering strategy that employs targeted prompts to guide reasoning and suppress redundant text generation.\\

        rStar2-Agent~\cite{shang2025rstar2}~\href{https://github.com/microsoft/rStar}{[code]} & SFT+RL & Mathematics & Introduce an efficient RL infrastructure with a tailored GRPO-RoC strategy, enabling a powerful agentic reasoning model.\\

        MedAgentGym~\cite{xu2025medagentgym}~\href{https://github.com/wshi83/MedAgentGym}{[code]} & SFT+RL & Healthcare & Advance coding-based medical reasoning by constructing a training environment that spans diverse biomedical scenarios.\\

         ML-Agent~\cite{liu2025ml}~\href{https://github.com/MASWorks/ML-Agent}{[code]} & SFT+RL & Machine Learning & Pioneers agentic machine learning engineering through online reinforcement learning in interactive environments.\\

    \midrule
       
        \rowcolor{red!5}
        \textbf{Agentic Visual Processing} & \textbf{Fine-tuning} &  \textbf{Processing Type} & \textbf{Contribution} \\
        \midrule
        
       DeepEyes~\cite{zheng2025deepeyes}~\href{https://github.com/Visual-Agent/DeepEyes}{[code]} & RL & Cropping  & Introduce a tool-use-oriented data selection mechanism and reward strategy to foster ``thinking with images'' capabilities. \\ 
       Ground-R1~\cite{cao2025ground} & RL & Cropping & Propose a reinforcement learning framework that Present scalable grounded visual reasoning without costly annotations.\\
       Active-O3~\cite{zhu2025active}~\href{https://github.com/aim-uofa/Active-o3}{[code]} & RL &  Cropping & Propose an RL framework that equips MLLMs with efficient active perception capabilities for tasks like small-object grounding.\\
       Chain-of-Focus~\cite{zhang2025chain}~\href{https://github.com/xtong-zhang/Chain-of-Focus}{[code]} & SFT+RL & Cropping & Enable MLLMs to perform adaptive region focusing and zooming through a two-stage training pipeline.\\
       Pixel-Reasoner~\cite{su2025pixel}~\href{https://github.com/TIGER-AI-Lab/Pixel-Reasoner}{[code]} & SFT+RL & Cropping & Propose pixel-space reasoning, a novel framework that equips MLLMs with visual operations (e.g., zoom-in, select-frame). \\
       VLM-R\textsuperscript{3}~\cite{jiang2025vlm} & SFT+RL & Cropping & Equip MLLMs with region recognition and reasoning capabilities via Region-Conditioned Reinforcement Policy Optimization. \\
       
       Mini-o3~\cite{lai2025mini-o3}~\href{https://github.com/Mini-o3/Mini-o3}{[code]} & SFT+RL & Cropping & Enable deep multi-turn reasoning with tool interactions, and achieves leading performance on complex visual search tasks. \\
       
       OpenThinkIMG~\cite{su2025openthinkimg}~\href{https://github.com/zhaochen0110/OpenThinkIMG}{[code]} & SFT+RL & Manipulation & Build a tool-augmented agentic MLLM with adaptive tool-use capabilities for complex chart reasoning tasks.\\
       Thyme~\cite{zhang2025thyme}~\href{https://github.com/yfzhang114/Thyme}{[code]} & SFT+RL & Manipulation & Present MLLMs to autonomously generate and execute image processing and computational code.\\
       VILASR~\cite{wu2025reinforcing}~\href{https://github.com/AntResearchNLP/ViLaSR}{[code]} & SFT+RL & Manipulation & Equip MLLMs with elementary drawing operations (e.g., bounding boxes, auxiliary lines) to enhance spatial reasoning.\\
       
       REVPT~\cite{zhou2025reinforced}~\href{https://github.com/ls-kelvin/REVPT}{[code]} & SFT+RL & Manipulation &  Enhance MLLMs' visual perception and reasoning by training them to dynamically leverage a suite of specialized visual tools. \\
       
       VPRL~\cite{xu2025visual}~\href{https://github.com/yix8/VisualPlanning}{[code]} & RL & Generation & Propose visual planning that replaces text-based reasoning with coherent image sequences generated by a large vision model. \\

    \bottomrule[1pt]
    \end{tabular}
    }
    \label{tab: 2 agentic external tool invocation}
\end{table*}

\item \textbf{Reasoning-driven memory.}
Building on these foundations, recent research has advanced toward reasoning-driven memory systems that autonomously store, update, and utilize memory in a more dynamic and task-driven manner.
A-Mem~\cite{xu2025a-mem} introduces an agentic memory framework inspired by the Zettelkasten method, allowing LLM agents to dynamically organize and evolve interconnected memory nodes for more adaptive, context-aware reasoning.
Mem0~\cite{chhikara2025mem0} proposes a scalable memory-centric architecture that dynamically manages salient information, with a graph-based variant to capture relational structures, yielding superior long-term conversational coherence.
MemTool~\cite{lumer2025memtool} focuses on short-term memory, enabling agents to dynamically manage tools or MCP server contexts across multi-turn conversations; it provides Autonomous, Workflow, and Hybrid modes with distinct trade-offs between efficiency and accuracy.  
More recently, Memory-R1~\cite{yan2025memoryr1} introduces an RL-based framework for adaptive external memory management. It employs a Memory Manager that learns structured operations (e.g., add, update, delete, noop) and an Answer Agent that retrieves and reasons over relevant entries, enabling continuous and flexible memory usage beyond static, rule-based approaches.

\end{itemize}

Despite these advances, most work on agentic memory remains text-centric, leaving a notable gap in multimodal agentic memory management for future research.

\subsection{Agentic External Tool Invocation}
\textit{``A good tool improves the way you work. A great tool improves the way you think.''~~~~~~~~~~~~~~~~~~~~~~~~\qquad\qquad--- Jeff Duntemann}

While internal intelligence equips agentic MLLMs with the ability to reason, reflect and memory, their capabilities remain intrinsically limited to the knowledge encoded in the model parameters.  
A natural strategy to overcome this limitation is to augment MLLMs with the ability to use external tools for problem solving.
Early approaches~\cite{jiang2024mmsearch, wang2025mllmtool, gao2024mm_agent_tuning} relied on prompt engineering to passively trigger tool use, but such methods lack the flexibility and adaptability required for novel tasks.  
Recent advances in agentic MLLMs have shifted this paradigm by integrating tool invocation into the reasoning process, enabling models to incorporate external tools into step-by-step reasoning and to autonomously determine when, and which tools to employ.
To this end, in this section, we review how agentic MLLMs learn to reason with external tools, categorized by different tool types, including information searching, code execution, and visual processing.
A summary of agentic external tool invocation method is presented in Table~\ref{tab: 2 agentic external tool invocation}.
\label{Sec: 4.2 external tool}

\subsubsection{Agentic Search for Information Retrieval}
\label{sec: 4.2.1 agentic search}
In today’s rapidly evolving information landscape, a pressing requirement for intelligent systems is the ability to stay current with emerging knowledge.  
However, once training is complete, the knowledge space of MLLMs becomes fixed and they cannot directly handle newly emerging events or rapidly changing domains.
For example, GPT-5~\cite{GPT-5}, released in August 2025, only retains knowledge up to June 2024, leaving it unable to address subsequent developments.
To overcome this limitation, researchers have proposed augmenting MLLMs with web search integration~\cite{nakano2021webgpt, zhang2024vsa, hu2023avis} or Retrieval-Augmented Generation (RAG)~\cite{chen2022murag, hu2023reveal}, enabling access to external knowledge sources such as the Internet or specialized databases.
This integration extends their capabilities beyond static parametric knowledge and enhances adaptability to dynamic real-world contexts.

\textbf{Search Agent.}
Traditional MLLM search agents~\cite{wang2025mllmtool, jiang2024mmsearch} often pre-define a sequential pipeline to execute search instructions and retrieve external knowledge for problem solving.
For example, when presented with an up-to-date question, the agent system first reformulates the query and submits it to a search engine, then reranks the retrieved results, and finally prompts the MLLM to synthesize the information into a coherent answer for the user.
Representative agent systems such as MMSearch~\cite{jiang2024mmsearch} and MindSearch~\cite{chen2024mindsearch} use this paradigm, proposing structured pipeline designs to enable access to external knowledge.

\textbf{Agentic Search.}
Agentic search leverages end-to-end reinforcement learning to equip MLLMs with the autonomy to decide both when to search and what to search for.
By embedding search directly into the reasoning process, this paradigm reduces redundant queries and enables retrieval that aligns more coherently with multi-turn interactions.
Training an agentic search model typically begins with curating up-to-date or knowledge-intensive questions that require external resources to answer, and constructing corresponding question–answer pairs.
Recent studies have proposed diverse strategies for building such datasets~\cite{geng2025webwatcher, liu2025visualarft, tao2025webshaper, wu2025mmsearchr1, su2025agenticCPT, wu2025webwalker}, including reverse engineering, graph-based synthesis, and formalized task modeling.
Based on these data, reinforcement learning with tailored reward functions \cite{liu2025visualarft, jin2025searchr1, wu2025mmsearchr1, geng2025webwatcher} is then employed to incentivize the model’s ability to conduct adaptive and contextually appropriate search.

Pioneering works focus on searching textual corpora.
For instance, Search-R1~\cite{wu2025mmsearchr1} integrates search into LLM reasoning with token-masked retrieval, interleaved multi-turn reasoning, and outcome-based rewards to stabilize RL training and enhance complex task solving.
Search-o1~\cite{li2025search-o1} introduces a framework that integrates agentic search into o1-like reasoning, enabling LLMs to retrieve and refine external knowledge on demand while preserving logical flow.
Moreover, Visual-ARFT~\cite{liu2025visualarft} augments multimodal understanding by integrating text search.

Building on this foundation, subsequent studies extend agentic search to multimodal information retrieval.
This is achieved via custom multimodal search frameworks~\cite{wang2025vrag-rl} or specialized engines such as Google SerpApi\footnote{SerpApi: \url{https://serpapi.com/}}.
VRAG-RL~\cite{wang2025vrag-rl} defines a visual action space with cropping and scaling for coarse-to-fine information gathering, reinforced by a reward combining query rewriting and retrieval accuracy. 
MMSearch-R1~\cite{wu2025mmsearchr1} integrates both image and text search tools, leveraging cold-start training and RL to teach models when and how to invoke each tool, guided by outcome-based rewards with search penalties. 
WebWatcher~\cite{geng2025webwatcher} further advances this line by systematically constructing search-dependent data and introducing unified special tokens to coordinate image and text search engines.

\begin{table*}[!ht]
    \setlength\tabcolsep{6pt}
    \centering
    \caption{Summary of agentic environment interaction, grouped into two categories: virtual and physical.
    }
    \resizebox{0.98\linewidth}{!}{
    \begin{tabular}{lccp{16cm}}
    \toprule[1pt]
\rowcolor{red!5}
        \textbf{Agentic Virtual Embodiment} & \textbf{Fine-tuning} &  \textbf{Learning Type} & \textbf{Contribution} \\
    \midrule
        AGUVIS~\cite{xu2024aguvis}~\href{https://github.com/xlang-ai/aguvis}{[code]} &  SFT & Offline & Integrate structured reasoning and operates autonomously as a unified vision-based GUI agent. \\
        InfiGUIAgent~\cite{liu2025infiguiagent}~\href{https://github.com/InfiXAI/InfiGUIAgent}{[code]} & SFT  & Offline & Cultivate native hierarchical and expectation-reflection reasoning skills to enhance multi-step GUI automation. \\
        TongUI~\cite{zhang2025tongui}~\href{https://github.com/TongUI-agent/TongUI-agent?tab=readme-ov-file}{[code]} & SFT & Offline &  Mitigate data scarcity for GUI agents by automatically generating the GUI-Net-1M dataset from multimodal web tutorials. \\

        UI-R1~\cite{lu2025uir1}~\href{https://github.com/lll6gg/UI-R1}{[code]}
        & RL  & Offline & Enable MLLMs to achieve significant accuracy improvements in GUI action prediction with exceptional data efficiency.\\
        
        GUI-R1 ~\cite{luo2025guir1}~\href{https://github.com/ritzz-ai/GUI-R1}{[code]} & RL & Offline &Leverage unified action space modeling and policy optimization to dramatically enhance the generalization and data efficiency. \\
        InfiGUI-R1~\cite{liu2025infiguir1}~\href{https://github.com/InfiXAI/InfiGUI-R1}{[code]} & RL & Offline & Introduce the Actor2Reasoner framework that transforms reactive GUI agents into deliberative reasoners.\\
        ComfyUI-R1~\cite{xu2025comfyui}~\href{https://github.com/AIDC-AI/ComfyUI-Copilot}{[code]} & SFT+RL & Offline &Propose a specialized reasoning model that achieves automated workflow generation through a two-stage training framework. \\

        GUI-Reflection~\cite{wu2025gui-reflection}~\href{https://github.com/penghao-wu/GUI_Reflection}{[code]} & Pretraining+SFT+RL & Online &Develop self-correction capabilities in GUI agents through automated reflection data generation and iterative online tuning. \\
        
        ZeroGUI~\cite{yang2025zerogui}~\href{https://github.com/OpenGVLab/ZeroGUI}{[code]}
        & RL & Online & Eliminate human annotation costs by automating task generation and reward estimation through MLLMs. \\
        WebAgent-R1~\cite{wei2025webagent}~\href{https://github.com/weizhepei/WebAgent-R1}{[code]}
        & RL & Online & Achieve strong gains in multi-turn web interactions via  asynchronous trajectory generation and binary reward optimization.\\
        UI-TARS~\cite{qin2025ui}~\href{https://github.com/bytedance/UI-TARS}{[code]}
        & Pretraining+SFT+RL & Online & Integrat innovations in screenshot perception, unified action modeling, deliberate reasoning and iterative self-improvement.\\
        
        UI-TARS-2~\cite{wang2025ui}~\href{https://github.com/bytedance/ui-tars}{[code]}
        & Pretraining+SFT+RL & Online & A systematic framework addressing data scalability, multi-turn RL stability, hybrid environment and unified sandbox platform. \\

    \midrule
\rowcolor{red!5}
        \textbf{Agentic Physical Embodiment} & \textbf{Fine-tuning} &  \textbf{Task Type} & \textbf{Contribution} \\
        \midrule

        ALP~\cite{liang2023alp}~\href{https://github.com/xinranliang/alp}{[code]} & RL & Perception &  Combine action-aware representation learning with active environmental exploration to learn robust visual representations. \\
        EAR~\cite{fan2024evidential} & RL & Perception & Model visual exploration as sequential evidence gathering with an uncertainty-aware reward for open-world environments. \\
        Wu et al.~\cite{wu2025reinforced}~\href{https://github.com/mail-taii/Reinforced-Reasoning-for-Embodied-Planning}{[code]} & SFT+RL & Planning &Incorporate R1-style reasoning to advance embodied planning performance and generalization in interactive environments. \\
        Embodied Planner-R1~\cite{fei2025unleashing}~\href{https://github.com/OpenMOSS/Embodied-Planner-R1}{[code]} & RL & Planning & Introduce an RL framework with sparse completion rewards and interactive policy optimization for embodied planning. \\

        OctoNav~\cite{gao2025octonav}~\href{https://github.com/buaa-colalab/OctoNav-R1}{[code]} & SFT+RL & Navigation & Construct a large-scale benchmark and a unified framework with think-before-action capability for generalist navigation agents.\\ 
        VLN-R1~\cite{qi2025vln}~\href{https://github.com/Qi-Zhangyang/GPT4Scene-and-VLN-R1}{[code]} & SFT+RL & Navigation & Propose a GRPO-based RL method with time-decayed rewards for continuous embodied navigation.\\
        Nav-R1~\cite{liu2025navr1}~\href{https://github.com/AIGeeksGroup/Nav-R1}{[code]} &  SFT+RL & Navigation & Decouple high-level planning from low-latency control and enables coherent yet highly responsive navigation. \\ 
        VLP~\cite{liu2025vlp}~\href{https://vlpref.github.io/}{[code]} & RL &Manipulation  & Advance a new approach to embodied manipulation via a language-conditioned preference feedback framework. \\
        ManipLVM-R1~\cite{song2025maniplvm} & RL & Manipulation & Develop an RL framework with two tailored reward functions for spatial perception and trajectory matching. \\
        Embodied-R1~\cite{yuan2025embodied}~\href{https://github.com/pickxiguapi/Embodied-R1}{[code]} & RL & Manipulation & Bridge the robotics perception-action gap with a pointing-centric representation and an RL-based training strategy. \\

    \bottomrule[1pt]
    \end{tabular}
    }
    \label{tab: 3 agentic enviroment interaction}
\end{table*}

\subsubsection{Agentic Coding for Complex Computations}
\label{sec: 4.2.2 agentic code}
While MLLMs have demonstrated remarkable capabilities in cross-modal vision-language tasks, they remain inherently limited in tasks requiring rigorous program synthesis, precise mathematical computation, and structured symbolic reasoning. A key development in overcoming these challenges is the emergence of agentic MLLMs, which autonomously plan, generate, and refine code-based actions through iterative program reasoning and dynamic tool utilization. 
Guided by the primary application domains of agentic coding, this section surveys these synergistic advances by categorizing recent work into three key areas: \textbf{program engineering}, \textbf{mathematical reasoning}, and \textbf{other domain-specific applications}.

\textbf{Program Engineering.} Recent research has extensively explored methods for taming LLMs to function as capable coding assistants~\cite{chen2021evaluating,li2022competition,guo2024deepseek}. The integration of RL has further augmented these capabilities, allowing for self-improvement in both code generation and execution accuracy~\cite{le2022coderl,shojaee2023execution,chen2025acereason}. One line of work utilizes outcome-based rewards, such as code execution and test case results, as direct training signals~\cite{chen2025acereason,feng2025towards}. In contrast, another strand of research introduces denser, process-oriented rewards that provide stepwise feedback on aspects such as code snippets and intermediate reasoning, thereby offering finer-grained guidance during training~\cite{dou2024stepcoder,ye2025process,fan2025posterior}. Subsequent studies have expanded these efforts, broadening the scope of agentic coding to include iterative code refinement through multi-turn interactions~\cite{chen2025r1}, co-evolution of code generators and unit testers to improve robustness~\cite{wang2025co}, and the application in advanced software engineering tasks~\cite{wang2024repogenreflex,lin2025r1}.

\textbf{Mathematical Reasoning.} Numerous studies have integrated external tools, such as computational libraries and symbolic solvers, directly into the reasoning process, a methodology now commonly termed tool-integrated reasoning. This enables models to dynamically execute code and obtain reliable numerical and symbolic solutions, significantly improving performance in complex reasoning tasks like mathematical problem-solving. Specifically, early efforts focus on building high-quality reasoning trajectories that interleave natural language reasoning with code execution, thereby stimulating the model’s capacity to autonomously generate and execute code~\cite{gou2023tora,wang2023mathcoder,guan2025rstar}. As a pioneering effort, ToRA~\cite{gou2023tora} first prompts advanced LLMs like GPT-4 to synthesize high-quality reasoning trajectories with tool calls for imitation learning. Subsequently, an output space shaping strategy is employed to augment the dataset with the initial model's self-generated correct trajectories and its errors after teacher model correction. A final SFT phase on this enriched data further enhances the model's capabilities in leveraging external tools and generating code to solve complex mathematical problems. 

Fueled by recent advances in large reasoning models, leveraging RL to autonomously integrate code generation into text-centric long CoT reasoning is an emerging research trend~\cite{li2025torl,feng2025retool}. For instance, ToRL~\cite{li2025torl} employs a pure RL strategy to promote code-integrated reasoning, while ReTool~\cite{feng2025retool} further enhances long-form capabilities through an outcome-driven RL framework that supports multi-turn code execution. Subsequent research has placed greater emphasis on balancing accuracy and efficiency in models that actively employ code generation for reasoning. In this vein, OTC~\cite{wang2025otc} introduces an Optimal Tool Call-controlled Policy Optimization that incentivizes models to solve tasks correctly using minimal tool interactions. CoRT~\cite{li2025cort} pinpoints two primary sources of inefficiency: first, a delay in code computation caused by a default to textual CoT reasoning prior to code generation; and second, a distrust in code results, which triggers superfluous manual verification of the execution outputs. To address these challenges, CoRT introduces a hint-engineering strategy that inserts strategic prompts to steer the reasoning trajectory, thereby avoiding the overhead of futile textual reasoning.

\textbf{Other domain-specific applications.} Beyond the above-mentioned advancements, recent research has successfully extended agentic coding techniques to a variety of other domains, e.g., healthcare~\cite{xu2025medagentgym} and machine learning~\cite{liu2025ml}. These cross-disciplinary efforts demonstrate the remarkable adaptability and impact of agentic coding, highlighting its potential to transform complex decision-making processes and operational workflows across diverse sectors.

\subsubsection{Agentic Visual Processing for Thinking with Image}
\label{sec: 4.2.3 agentic data process}
Recent advances demonstrate a paradigm shift in large reasoning models from text-centric approaches towards integrated multimodal reasoning, which jointly interleaves textual and visual information. This evolution is often driven by the agentic invocation of tools or functions, enabling a form of ``thinking with images''~\cite{o3}. Based on their distinct approaches to image processing, we can roughly categorize the evolution into three main phases: thinking with \textbf{cropped} images, thinking with \textbf{manipulated} images, and thinking with \textbf{generated} images.

\textbf{Thinking with cropped images:} As the early open-source initiative of its kind, DeepEyes~\cite{zheng2025deepeyes} effectively integrates visual information into textual chain-of-thought reasoning by leveraging the model's inherent grounding capabilities, augmented with cropping and zoom-in functions. The training framework relies exclusively on reinforcement learning (i.e., GRPO) with tailored reward functions, eliminating the need for cold-start SFT. Concurrent works such as Ground-R1~\cite{cao2025ground} and Active-O3~\cite{zhu2025active} implement similar RL-driven concepts, differing only marginally in their use of training data and reward design. Another line of research, exemplified by methods such as Chain-of-Focus~\cite{zhang2025chain}, Pixel-Reasoner~\cite{su2025pixel}, and VLM-R\textsuperscript{3}~\cite{jiang2025vlm}, employs cold-start SFT to equip models with multimodal reasoning strategies and structured output formats in advance, thereby alleviating the burden on subsequent reinforcement learning. These approaches sample their initial training data from existing datasets such as VisCoT~\cite{shao2024visual}, or leverage GPT-4o to curate examples based on image collections like SA-1B~\cite{kirillov2023segment}. Remarkably, the latest Mini-o3~\cite{lai2025mini-o3} achieves deep multi-turn exploration with tool interactions through specialized dataset construction, diverse trajectory collection, and innovative over-turn masking strategies, leading to state-of-the-art performance on challenging visual search tasks.

\textbf{Thinking with manipulated images:} Beyond fundamental operations such as cropping and zooming, more advanced approaches endow models with enhanced capabilities for active image manipulation. OpenThinkIMG~\cite{su2025openthinkimg} builds a tool-augmented, agentic MLLM with adaptive tool-use capabilities for complex chart reasoning tasks. The toolset encompasses both basic operations (e.g., crop, zoom-in, and draw) and powerful external models including SAM~\cite{kirillov2023segment} and GroundingDino~\cite{liu2024grounding}. Thyme~\cite{zhang2025thyme} utilizes agentic code generation to perform autonomous image editing (e.g., cropping, rotation, contrast enhancement) and mathematical computations, within its reasoning process. This method uses a two-stage SFT and RL training paradigm and introduces GRPO with Adaptive Temperature Sampling (GRPO-ATS), which decouples text and code sampling temperatures to ensure high-fidelity code generation. VILASR~\cite{wu2025reinforcing} extends this concept to spatial intelligence, enabling the model to edit images or video frames by drawing additional bounding boxes or auxiliary lines. Experiments across multiple benchmarks confirm that this method consistently boosts spatial reasoning performance. Furthermore, ReVPT~\cite{zhou2025reinforced} incorporates a comprehensive visual toolkit, including depth estimation, zoom in, object detection, and edge detection. Empowered by cold-start SFT and RL training, it demonstrates significantly enhanced visual perception, setting a new state-of-the-art on spatial reasoning and image understanding benchmarks.

\textbf{Thinking with generated images:} Recently, a growing number of efforts extend reinforcement learning to image generation, leveraging it to unlock MLLM reasoning for creating high-fidelity images that are better aligned with human instructions~\cite{duan2025got,guo2025can,tong2025delving}. In parallel, another line of research explores active image generation for enhanced visual understanding. For instance, the VPRL~\cite{xu2025visual} method employs reinforcement learning to endow large vision models with visual chain-of-thought reasoning capabilities. By generating a sequence of images that provides coherent visual cues, these models achieve significant performance gains in visual planning tasks.

\subsection{Agentic Environment Interaction} 
\label{Sec: 4.3 environment interaction}

Beyond reasoning and tool utilization, agentic environment interaction represents the stage where MLLMs transcend static query-response paradigms and begin engaging with their surroundings.
Through continuous virtual or physical interaction (i.e., executing actions, perceiving environmental changes, and integrating feedback), agentic MLLMs dynamically adjust their strategies in response to evolving contexts, enabling them to pursue long-term goals, adapt in real time, and align their behaviors with the surrounding environment.
A summary of agentic environment interaction method is presented in Table~\ref{tab: 3 agentic enviroment interaction}.

\subsubsection{Agentic Virtual Environment Interaction}
\label{sec: 4.3.1 agentic Virtual Embodiment}
Recent years have witnessed significant advances in agentic MLLMs capable of performing complex tasks through graphical user interfaces (GUIs). 
These GUI agents, which enable autonomous interaction with digital environments, have evolved into increasingly sophisticated systems that leverage learning-based approaches to generalize across diverse applications and platforms. In this section, we categorize these systems based on their learning mechanisms: one that \textbf{learn from pre-collected GUI demonstration trajectories}, and another that \textbf{learn directly through interaction within dynamic GUI environments}. We systematically examine both categories, highlighting their representative methods, key strengths, and inherent limitations.

\textbf{Learning from offline demonstration trajectories:}
AGUVIS~\cite{xu2024aguvis} introduces a large-scale GUI trajectory dataset and a two-stage training framework that decouples visual grounding from high-level planning, establishing state-of-the-art performance across offline and online GUI benchmarks. InfiGUIAgent~\cite{liu2025infiguiagent} also adopts a two-stage SFT workflow that first instills core GUI grounding skills, then enhances reasoning and reflection capabilities using synthesized data. TongUI~\cite{zhang2025tongui} addresses the critical bottleneck of limited training data for generalized GUI agents by automatically constructing a large-scale, multimodal dataset, termed GUI-Net-1M, from crawled web tutorials. By fine-tuning the Qwen2.5-VL~\cite{bai2025qwen2.5vl} models on this dataset, the resulting TongUI agent demonstrates a substantial performance gain on standard grounding and navigation benchmarks, validating the framework's effectiveness and the utility of the newly created resource. 

Despite technical progress, the conventional SFT training paradigm exhibits a strong dependency on massive, curated datasets and hinders model generalization in unseen environments. To address this limitation, significant research efforts are devoted to integrating RL into GUI-based tasks. UI-R1~\cite{lu2025uir1} first proposes an RL framework that significantly enhances GUI action prediction through policy optimization with novel rule-based action-level rewards. This approach demonstrates remarkable data efficiency, achieving substantial accuracy gains on both in-domain and out-of-domain mobile GUI tasks using only 136 training examples. GUI-R1~\cite{luo2025guir1} further boosts the real-world problem-solving capabilities of MLLMs via unified action space modeling and policy optimization, achieving state-of-the-art performance across multiple platforms in a highly data-efficient manner.
InfiGUI-R1~\cite{liu2025infiguir1} posits that advancing GUI agents requires a fundamental shift from reactive actors to deliberative reasoners and introduces the Actor2Reasoner framework, a novel two-stage training methodology. Specifically, it first injects explicit spatial reasoning capabilities through distillation and then enhances deliberation via reinforcement learning with sub-goal guidance and error recovery scenarios construction, yielding superior cross-platform performance. ComfyUI-R1~\cite{xu2025comfyui} presents a two-stage training framework that achieves cutting-edge automated workflow generation. The framework first adapts a model to the ComfyUI domain via CoT fine-tuning and then enhances its reasoning through RL with a novel rule-metric hybrid reward. 


As a common and stable approach, offline learning from demonstration trajectories provides a solid foundation for GUI automation. However, models trained this way lack the robustness to handle real-world challenges such as unexpected events and execution errors. To bridge this gap, research has pivoted to training models via direct online interaction in dynamic GUI environments.

\textbf{Learning from online GUI Environments:} 
GUI-Reflection~\cite{wu2025gui-reflection} significantly enhances the self-reflection and error recovery capabilities of GUI automation by introducing automated data generation and iterative online tuning. This creates a new paradigm for building robust GUI agents capable of autonomous operation and error correction without the need for human annotation. ZeroGUI~\cite{yang2025zerogui} also introduces a scalable online learning framework that eliminates the dependency on human annotations by automating both task generation and reward estimation through MLLMs. Leveraging the tailored two-stage RL process, the GUI agent enables continuous adaptation to dynamic GUI environments via autonomous interaction and self-improvement. WebAgent-R1~\cite{wei2025webagent} presents an end-to-end RL framework that addresses the challenges of multi-turn decision-making in dynamic web environments by learning directly from binary task-completion rewards. UI-TARS~\cite{qin2025ui} is a novel end-to-end native GUI agent that achieves unprecedented performance by integrating four key innovations: enhanced perception with large-scale GUI data, unified cross-platform action modeling, deliberate System-2 reasoning, and iterative self-improvement through reflective online trace tuning. UI-TARS-2~\cite{wang2025ui} further features a next-generation native GUI agent. Through a systematic methodology that incorporates scalable data generation, stabilized multi-turn reinforcement learning, hybrid environment integration, and a unified sandbox platform, it achieves state-of-the-art performance on both standard GUI benchmarks and complex game environments.

\subsubsection{Agentic Physical Environment Interaction}
\label{sec: 4.3.1 agentic Physical Embodiment}
Embodied AI distinguishes itself by creating autonomous agents capable of active perception, deliberate reasoning, and physical interaction within real-world environments. This paradigm aligns closely with agentic MLLMs, as both transcend passive comprehension to exhibit goal-driven, intentional behavior. By integrating sensing, planning, and acting in a closed-loop system, embodied agents underscore a pivotal shift toward models that not only interpret context but also engage with it dynamically. In this section, we explore the core capabilities that enable autonomous operation and structure our discussion into four key areas: embodied \textbf{perception}, \textbf{planning}, \textbf{navigation} and \textbf{manipulation}.

\textbf{Embodied Perception:}
A substantial body of research is dedicated to embodied perception, a foundational concept in embodied AI wherein an agent acquires information through active, deliberate environmental exploration to guide its subsequent actions~\cite{das2018neural,chaplot2020learning,jayaraman2018learning}. For instance, ALP~\cite{liang2023alp} proposes an embodied learning framework that integrates action-aware representation learning with active environmental exploration to learn more robust and generalizable visual representations compared to static dataset training approaches. EAR~\cite{fan2024evidential} proposes an uncertainty-aware active recognition framework that models visual exploration as sequential evidence gathering with theoretical uncertainty quantification and reliable prediction. Incentivized by a tailored open-world reward function, this framework demonstrates superior performance in both recognition accuracy and robustness.

\textbf{Embodied Planning:}
Building upon the actively perceptual understanding, embodied planning requires the agent to formulate a sequence of actionable steps or decisions to achieve a long-horizon goal, effectively bridging perception with concrete execution. To advance embodied planning, Wu et al.~\cite{wu2025reinforced} propose a novel reinforcement fine-tuning framework that integrates R1-style reasoning with structured decision-making priors. Through SFT and rule-based generalized reinforced preference optimization, this method significantly enhances embodied planning performance and generalization in interactive environments. Embodied Planner-R1~\cite{fei2025unleashing} also incorporates RL into planning. Leveraging sparse outcome rewards and interactive policy optimization, it demonstrates superior completion ratios and robustness across multiple benchmarks.

\textbf{Embodied Navigation:}
As a core instantiation of embodied planning, embodied navigation focuses on the agent's ability to traverse through physical or simulated spaces by leveraging its perceptual inputs and planned path to reach a specified destination. Towards the goal of generalist navigation agents, OctoNav~\cite{gao2025octonav} unifies multiple navigation tasks with a new benchmark (OctoNav-Bench) and method (OctoNav-R1). Leveraging a hybrid training paradigm, OctoNav-R1 operates in a ``think-before-act'' mode, demonstrating impressive navigation performance. VLN-R1~\cite{qi2025vln} introduces an end-to-end framework that enables continuous vision-language navigation through direct egocentric video-to-action translation, combining an innovative long-short memory approach and time-decayed reward mechanisms to achieve strong benchmark performance through data-efficient reinforcement learning. Nav-R1~\cite{liu2025navr1} further advances embodied navigation with its Fast-in-Slow reasoning framework. This dual system separates high-level semantic planning from time-critical reactive control, enabling robust and coherent navigation in dynamic environments without sacrificing real-time results.

\textbf{Embodied Manipulation:}
Extending beyond navigation, embodied manipulation involves the agent interacting with and altering its environment through physical actions, thereby completing embodied tasks that require both motion and interaction with objects. Specifically, VLP~\cite{liu2025vlp} addresses the annotation bottleneck in preference-based RL via a well-designed vision-language framework that autonomously generates language-conditioned preferences for embodied manipulation tasks, facilitating scalable policy learning and robust generalization to novel instructions and tasks. ManipLVM-R1~\cite{song2025maniplvm} eliminates human annotation dependency through a reinforcement learning framework with two specialized rewards: Affordance Perception Reward for spatial interaction and Trajectory Match Reward for physical plausibility. Experiments show it achieves higher performance gains and better generalization with reduced training data. Embodied-R1~\cite{yuan2025embodied} addresses the challenging ``seeing-to-doing'' gap in robotics by introducing pointing as a unified intermediate representation. Through a two-stage reinforced fine-tuning framework, it achieves exceptional zero-shot generalization, offering valuable insights for the broader embodied AI community. 

\section{Training \& Evaluation}
\label{sec: 5 training and evaluation}
In order to develop and assess agentic MLLMs, three core components are indispensable:
\textbf{training frameworks} that provide the algorithmic and optimization infrastructure,
\textbf{training datasets} that foster agentic cross-modal alignment and robust generalization,
and \textbf{evaluation datasets} that measure the capabilities of agentic MLLMs.
Therefore, this section surveys the landscape of open resources for agentic MLLMs across these three dimensions, helping the community to advance agentic research.

\subsection{Training Framework}
\begin{table*}[!ht]
\setlength\tabcolsep{16pt}
\centering
\caption{Summary of training framework for agentic CPT, SFT, and RL.}
\resizebox{0.75\linewidth}{!}{
\begin{tabular}{l l c l l}
\toprule[1pt]
\rowcolor{red!5}
\textbf{Framework} & \textbf{Link}  & \textbf{Type} & \textbf{Supports MLLM} & \textbf{Key Features} \\
\midrule
\rowcolor{blue!5} 
\multicolumn{5}{c}{\textbf{Agentic CPT/SFT Frameworks}} \\
\midrule
LLaMA-Factory~\cite{zheng2024llamafactory} & \href{https://github.com/hiyouga/LLaMA-Factory}{Code} & Agentic CPT/SFT & Yes & Easy, Various and Efficient Fine-tuning\\ 
MS-Swift~\cite{zhao2024swift} & \href{https://github.com/modelscope/ms-swift}{Code} & Agentic CPT/SFT & Yes & Scalable Lightweight Infrastructure \\
Megatron-LM~\cite{megatron-lm} & \href{https://github.com/NVIDIA/Megatron-LM}{Code} & Agentic CPT/SFT & Yes & GPU-optimized library \\
Unsloth~\cite{unsloth} & \href{https://github.com/unslothai/unsloth}{Code} & Agentic CPT/SFT & Yes &  Accurate, Accessible, Efficient\\
FireAct~\cite{chen2023fireact} & \href{https://github.com/anchen1011/FireAct}{Code} & Agentic CPT/SFT & No &  Language Agent Fine-tuning\\
AgenTuning~\cite{zeng2023agenttuning} & \href{https://github.com/THUDM/AgentTuning}{Code} & Agentic CPT/SFT & No &  Generalized Agent Abilities\\
LMFlow~\cite{diao2023lmflow} & \href{https://github.com/OptimalScale/LMFlow}{Code} & Agentic CPT/SFT & Yes & Extensible, Efficient, User-friendly, Open\\
\midrule
\rowcolor{blue!5} 
\multicolumn{5}{c}{\textbf{Standard RL Frameworks}} \\
\midrule
TRL~\cite{vonwerra2022trl} & \href{https://github.com/huggingface/trl}{Code} & RL & No & HuggingFace PPO/DPO Fine-tuning  \\
Open R1~\cite{openr1} & \href{https://github.com/huggingface/open-r1}{Code} & RL & No & DeepSeek-R1 Reproduction  \\
OpenRLHF~\cite{hu2024openrlhf} & \href{https://github.com/OpenRLHF/OpenRLHF}{Code} & RL & No & Comprehensive, Lightweight, Easy-to-use \\
Multimodal Open R1  & \href{https://github.com/EvolvingLMMs-Lab/open-r1-multimodal}{Code} & RL & Yes & Multimodal R1 Training\\
Logic-RL~\cite{xie2025logicrlunleashingllmreasoning} & \href{https://github.com/Unakar/Logic-RL}{Code} & RL & No & Rule-based RL Reasoning\\
EasyR1~\cite{zheng2025easyr1} & \href{https://github.com/hiyouga/EasyR1}{Code} & RL & Yes & Efficient Multi-modal RL  \\
Simple-R1~\cite{zeng2025simplerl} & \href{https://github.com/hkust-nlp/simpleRL-reason}{Code} & RL & No & Simple RL Reasoning  \\
Light-R1~\cite{wen2025light} & \href{https://github.com/Qihoo360/Light-R1}{Code} & RL & No & Curriculum SFT + RL  \\
R1-V~\cite{chen2025r1v} & \href{https://github.com/StarsfieldAI/R1-V}{Code} & RL & Yes & General VLM RL  \\
AReal~\cite{fu2025areal} & \href{https://github.com/inclusionAI/AReaL}{Code} & RL & No & Fully Asynchronous RL  \\
\midrule
\rowcolor{blue!5}
\multicolumn{5}{c}{\textbf{Agentic RL Frameworks}} \\
\midrule
verl & \href{https://github.com/volcengine/verl}{Code}  & Agentic RL & Yes & Flexible, Efficient RL library \\
RLFactory & \href{https://github.com/Simple-Efficient/RL-Factory}{Code} & Agentic RL & Yes &  Easy, Efficient Agentic Learning  \\
Visual-ARFT~\cite{liu2025visualagenticreinforcementfinetuning} & \href{https://github.com/Liuziyu77/Visual-RFT}{Code} & Agentic RL & Yes & Flexible Agentic LVLM   \\
rLLM~\cite{rllm2025} & \href{https://github.com/rllm-org/rllm}{Code} & Agentic RL & No & Customizable Agent Training  \\
Search-R1~\cite{jin2025search} & \href{https://github.com/PeterGriffinJin/Search-R1}{Code} & Agentic RL & No & LLM with Search Tool \\
MMSearch-R1~\cite{wu2025mmsearchr1} & \href{https://github.com/EvolvingLMMs-Lab/multimodal-search-r1}{Code} & Agentic RL & Yes & Multimodal Search Agent \\
Agent Lightning~\cite{luo2025agentlightningtrainai} & \href{https://github.com/microsoft/agent-lightning}{Code} & Agentic RL & No & Train-any-agent without Modifying \\
RAGEN~\cite{ragen} & \href{https://github.com/RAGEN-AI/RAGEN}{Code} & Agentic RL & No & RL + LLM + Agents \\
MARTI~\cite{marti2025} & \href{https://github.com/TsinghuaC3I/MARTI}{Code} & Agentic RL & No & Multi-agent RL \\
MiroRL~\cite{2025mirorl} & \href{https://github.com/MiroMindAI/MiroRL}{Code} & Agentic RL & No & Multi-turn MCP Tool \\
ROLL~\cite{wang2025reinforcement} & \href{https://github.com/alibaba/ROLL}{Code} & Agentic RL & No & User-friendly Large-scale RL \\
SkyRL~\cite{cao2025skyrl} & \href{https://github.com/NovaSky-AI/SkyRL}{Code} & Agentic RL & No & Modular Full-stack RL \\
AWorld~\cite{yu2025aworldorchestratingtrainingrecipe} & \href{https://github.com/inclusionAI/AWorld}{Code} & Agentic RL & No & Agent Self-improvement at Scale  \\
AgentFly~\cite{wang2025agentfly} & \href{https://github.com/Agent-One-Lab/AgentFly}{Code} & Agentic RL & Yes & Multi-turn, Async tool, Multimodal  \\
\bottomrule[1pt]
\end{tabular}
}
\label{tab:rl-frameworks}
\end{table*}
\label{sec: 5.1 framework}

In this section, we review open-source training frameworks that support agentic continual pre-training, supervised fine-tuning, and reinforcement learning. 
These frameworks provide code implementations and advanced training optimizations that facilitate efficient development of agentic MLLMs. 
A summary of training framework is shown in Table~\ref{tab:rl-frameworks}, with corresponding links for ease of access.

\textbf{Agentic CPT/SFT Frameworks.}
Llama-Factory~\cite{zheng2024llamafactory} is an open-source, user-friendly framework that provides efficient, extensible, and unified pipelines for fine-tuning large language models across diverse tasks and settings.
Ms-swift~\cite{zhao2024swift} is a versatile framework for training, aligning, and deploying large language and multi-modal models with advanced techniques.
unsloth~\cite{unsloth} is a cross-platform toolkit enabling efficient, exact-accuracy finetuning of diverse transformer models on standard NVIDIA GPUs without hardware changes.
FireAct~\cite{chen2023fireact} provides code, prompts, and datasets for fine-tuning language agents, along with model family descriptions for research use.
AgentTuning~\cite{zeng2023agenttuning} introduces instruction-tuning with agent trajectories, enhancing LLMs’ agent capabilities.
LMFlow~\cite{diao2023lmflow} is an extensible and user-friendly toolbox for efficient finetuning of large machine learning models.

\textbf{Standard RL Frameworks.}
verl~\cite{sheng2024hybridflow} is a flexible RL training library for large language models, implementing HybridFlow RLHF. 
TRL~\cite{vonwerra2022trl} provides a toolkit for post-training transformers via RL algorithms such as PPO and DPO. 
Open R1~\cite{openr1} is an open reproduction of DeepSeek-R1’s reasoning pipeline, democratizing chain-of-thought training. 
OpenRLHF~\cite{hu2024openrlhf} offers a scalable Ray-based RLHF framework supporting PPO and GRPO. 
Multimodal Open R1 adds multi-modal input support to the Open R1~\cite{openr1} pipeline. 
Logic-RL~\cite{xie2025logicrlunleashingllmreasoning} introduces rule-based RL to teach logical reasoning through strict reward shaping. 
EasyR1~\cite{zheng2025easyr1} is an efficient RL training framework supporting multimodality, achieving gains on reasoning benchmarks. 
Simple-R1~\cite{zeng2025simplerl} explores ``zero-start’’ RL training, showing even small models can benefit from RL on reasoning tasks. 
Light-R1~\cite{wen2025light} combines supervised fine-tuning, DPO, and RL to build reasoning models from scratch. 
R1-V~\cite{chen2025r1v} improves VLM reasoning at a very low cost, demonstrating strong generalization. 
AReaL~\cite{fu2025areal} is a fully asynchronous, open-source RL training system for large reasoning models that emphasizes reproducibility and accessibility for building AI agents.

\textbf{Agentic RL Frameworks.}
RLFactory is an agentic RL post-training framework that decouples environment setup from training and supports asynchronous tool-calling for faster agent learning. 
Visual-ARFT~\cite{liu2025visualagenticreinforcementfinetuning} equips open-source LVLMs with flexible agentic abilities for real-time web browsing and image manipulation, and introduces the MAT benchmark to evaluate multimodal search and coding skills.
The rLLM framework~\cite{rllm2025} provides abstractions to define custom language agents and environments, unifying inference and training with efficient scaling. 
Search-R1~\cite{jin2025search} trains LLMs to interleave reasoning with search engine calls, encouraging retrieval-based reinforcement learning. 
MMSearch-R1~\cite{wu2025mmsearchr1} enables multi-modal models to perform multi-turn real-world search with reinforcement. 
Agent Lightning~\cite{luo2025agentlightningtrainai} can train virtually any agent with RL while requiring minimal code changes. 
RAGEN~\cite{ragen} leverages RL to train LLM-based reasoning agents in stochastic environments, enabling self-evolution behaviors. 
MARTI~\cite{marti2025} combines centralized multi-agent interactions with distributed training, supporting scalable LLM collaboration.
MiroRL~\cite{2025mirorl} is the first RL framework enabling multi-turn MCP tool calls, offering agents seamless access to diverse tools while ensuring stable, efficient, and scalable training.
ROLL~\cite{wang2025reinforcement} is a unified and user-friendly RL library for large-scale LLM optimization. 
SkyRL~\cite{cao2025skyrl} is a modular full-stack RL library that integrates agent layers, training modules, and environments for multi-turn tasks.
AWorld~\cite{yu2025aworldorchestratingtrainingrecipe} enables large-scale agent self-improvement through continual learning from knowledge and experience.
AgentFly~\cite{wang2025agentfly} is an extensible RL framework for multi-turn, asynchronous, and multimodal agent training with easy tool and reward integration.

\subsection{Training dataset}
\begin{table*}[ht]
\setlength\tabcolsep{16pt}
\centering
\caption{Summary of datasets for \textbf{training} agentic MLLMs, where T, I, V, and A represent text, image, video, and audio.}
\resizebox{0.8\linewidth}{!}{
\begin{tabular}{l l l l l l l }
\toprule[1pt]
\rowcolor{red!5}
\textbf{Training Dataset} & \textbf{Link} & \textbf{Stage} &  \textbf{Type} & \textbf{Scope} & \textbf{Modality} & \textbf{Samples}  \\
\midrule
\rowcolor{blue!5}
\multicolumn{7}{c}{\textbf{Agentic Internal Intelligence}} \\
\midrule

MAVIS~\cite{zhang2024mavis} & \href{https://github.com/ZrrSkywalker/MAVIS}{Data} & SFT & Reasoning & Math & T, I & 834K \\
R$^3$V~\cite{r3v_reflection} & \href{https://github.com/njucckevin/MM-Self-Improve}{Data} & SFT & Reasoning, Reflection & Chart, Math & T, I & 5K \\
LLaVA-CoT-100k~\cite{xu2025llavacotletvisionlanguage} & \href{https://huggingface.co/datasets/Xkev/LLaVA-CoT-100k}{Data} & SFT & Reasoning &  Diverse & T, I & 100K \\
Mulberry-260K~\cite{yao2024mulberryempoweringmllmo1like}  & \href{https://huggingface.co/datasets/HuanjinYao/Mulberry-SFT}{Data}  &  SFT & Reasoning, Reflection & Diverse & T, I & 260K \\
Vision-R1-cold-200K~\cite{huang2025visionr1} & \href{https://huggingface.co/datasets/Osilly/Vision-R1-cold}{Data} &  SFT & Reasoning, Reflection &  Diverse  & T, I & 200K \\
R1-OneVision~\cite{yang2025r1onevision} & \href{https://huggingface.co/datasets/Fancy-MLLM/R1-Onevision}{Data} & SFT + RL & Reasoning & Diverse & T, I & 155K \\
MM-K12~\cite{meng2025mmekura} & \href{https://huggingface.co/datasets/FanqingM/MMK12}{Data} & RL & Reasoning &  Math & T, I & 15K \\
OpenVLThinker~\cite{deng2025openvlthinkerearlyexplorationcomplex} & \href{https://huggingface.co/collections/ydeng9/openvlthinker-v12-datasets-686f45e48d02e00b1585299e}{Data} & SFT + RL & Reasoning & Diverse & T, I & 12K \\
ThinkLite-VL~\cite{wang2025thinklitevl} & \href{https://huggingface.co/datasets/russwang/ThinkLite-VL-hard-11k}{Data} & RL & Reasoning & Diverse & T, I& 11K \\
Revisual-R1~\cite{revisual-r1} & \href{https://github.com/CSfufu/Revisual-R1}{Data} & SFT + RL & Reasoning & Diverse & T, I & 99K \\
GThinker-11k~\cite{zhan2025gthinker} & \href{https://huggingface.co/collections/JefferyZhan/gthinker-683e920eff706ead8fde3fc0}{Data} & SFT + RL & Reasoning & Diverse &  T, I & 11K \\

Video-R1~\cite{feng2025videor1} & \href{https://huggingface.co/datasets/Video-R1/Video-R1-data}{Data} & SFT + RL & Reasoning & Diverse & T, I, V & 425K  \\

MedTVT-QA~\cite{zhang2025medtvt} &\href{https://huggingface.co/datasets/kekeYeah/MedTVT-QA}{Data} & SFT + RL & Reasoning & Medical & T, I &  8K \\

WeThink~\cite{yang2025wethink} & \href{https://huggingface.co/datasets/yangjie-cv/WeThink_Multimodal_Reasoning_120K}{Data} & SFT + RL & Reasoning & Diverse  & T, I & 120K \\

AVQA-R1-6K~\cite{xing2025Echoink-r1} & \href{https://huggingface.co/datasets/harryhsing/AVQA-R1-6K}{Data} & RL & Reasoning & Diverse & T, I, A & 6K \\

Video-XL-pro~\cite{liu2025video-xl-pro} & \href{https://github.com/VectorSpaceLab/Video-XL/tree/main/Video-XL-Pro}{Data} & SFT& Memory & Dirvese & T, V & 3,000K \\

Long-VILA~\cite{chen2024longvila} & \href{https://github.com/NVlabs/VILA/tree/main/longvila}{Data} & SFT + RL & Memory & Diverse & T, V & 71K \\

\midrule
\rowcolor{blue!5}
\multicolumn{7}{c}{\textbf{Agentic External Tool Invocation}} \\
\midrule

Search-R1~\cite{jin2025searchr1}& \href{https://huggingface.co/collections/PeterJinGo/search-r1-67d1a021202731cb065740f5}{Data} & RL & Search & Multi-hop & T & 170K  \\

Search-o1~\cite{li2025search-o1} & \href{https://github.com/RUC-NLPIR/Search-o1}{Data} & RL & Search & Multi-hop & T & 1K\\

R1-Searcher~\cite{song2025r1-searcher} & \href{https://huggingface.co/datasets/XXsongLALA/RAG-RL-Hotpotqa-with-2wiki}{Data} & RL & Search & Multi-hop & T & 8K \\

FVQA~\cite{wu2025mmsearchr1} & \href{https://huggingface.co/datasets/lmms-lab/FVQA}{Data} & RL & Search & Multi-hop & T, I & 5K \\

MAT-Training~\cite{liu2025visualarft} & \href{https://huggingface.co/datasets/laolao77/MAT}{Data} & RL & Search, Code & Multi-hop, Code & T, I & 3K  \\

MathCoder~\cite{wang2023mathcoder} & \href{https://huggingface.co/datasets/MathLLMs/MathCodeInstruct}{Data} & SFT & Code & Math, Code & T & 80K  \\

ReTool~\cite{feng2025retool} & \href{https://huggingface.co/datasets/JoeYing/ReTool-SFT}{Data} & SFT & Code & Math, Code & T & 2K  \\

ToRL~\cite{li2025torl} & \href{https://github.com/GAIR-NLP/ToRL/tree/main/data/torl_data}{Data} & RL & Code & Math, Code & T & 28K  \\

rStar-Coder~\cite{liu2025rstar} & \href{https://huggingface.co/datasets/microsoft/rStar-Coder}{Data} & SFT + RL & Code & Math, Code & T & 580K  \\

DeepEyes~\cite{zheng2025deepeyes} & \href{https://huggingface.co/datasets/ChenShawn/DeepEyes-Datasets-47k}{Data} & RL & Visual Processing & Diverse & T, I & 47K  \\

Pixel-Reasoner~\cite{su2025pixel} & \href{https://huggingface.co/collections/TIGER-Lab/pixel-reasoner-682fe96ea946d10dda60d24e}{Data} & SFT + RL & Visual Processing & Diverse & T, I & 23K  \\

Chain-of-Focus~\cite{zhang2025chainoffocus} & \href{https://huggingface.co/datasets/xintongzhang/CoF-SFT-Data-5.4k}{Data} & SFT & Visual Processing & Diverse & T, I & 5K  \\

Mini-o3~\cite{lai2025mini-o3} & \href{https://huggingface.co/Mini-o3/datasets}{Data} & SFT + RL & Visual Processing & Diverse & T, I & 14K  \\

Thyme~\cite{zhang2025thyme} & \href{https://huggingface.co/collections/Kwai-Keye/thyme-689ebea74a628c3a9b7bd789}{Data} & SFT + RL & Visual Processing & Diverse & T, I & 401K  \\

\midrule
\rowcolor{blue!5}
\multicolumn{7}{c}{\textbf{Agentic Environment Interaction}} \\
\midrule
GUI-World~\cite{chen2024gui-world} & \href{https://huggingface.co/datasets/ONE-Lab/GUI-World}{Data} & SFT & Virtual & GUI & T, V & 12K\\
Show-UI~\cite{lin2025showui} & \href{https://huggingface.co/datasets/showlab/ShowUI-desktop}{Data} & SFT & Virtual & GUI & T, I & 8K \\

GUI-R1-3K~\cite{luo2025guir1}  & \href{https://huggingface.co/datasets/ritzzai/GUI-R1}{Data} & RL & Virtual & GUI & T, I & 3K \\
UI-R1~\cite{lu2025uir1} & \href{https://huggingface.co/datasets/LZXzju/UI-R1-3B-Train}{Data} & RL & Virtual & GUI & T, I & 136  \\
GUI-Reflection~\cite{wu2025gui-reflection} & \href{https://huggingface.co/collections/craigwu/gui-reflection-683c7fb964b44c0cca842290}{Data} & SFT & Virtual & GUI & T, I & 296K  \\

VLN-Ego~\cite{qi2025vln} & \href{https://huggingface.co/datasets/alexzyqi/VLN-Ego}{Data} &SFT + RL &Physical  & Navigation & T, V & 1.8M\\

InternData-N1~\cite{interndata_n1}  & \href{https://huggingface.co/datasets/InternRobotics/InternData-N1}{Data} & SFT &Physical & Navigation &T, V& 370K\\

VLA-IT~\cite{yang2025instructvla} & \href{https://huggingface.co/datasets/ShuaiYang03/VLA_Instruction_Tuning}{Data} & SFT &Physical & Manipulation & T, I & 650K\\

\bottomrule
\end{tabular}
}
\label{tab:agentic-MLLM-training-dataset}
\end{table*}
\label{sec: 5.2 training dataset}
In this section, we review publicly available training datasets that support the development of agentic capabilities, including internal intelligence, external tool invocation, and environment interaction.
Corresponding links are provided for easy access and practical use, as shown in Table~\ref{tab:agentic-MLLM-training-dataset}.

\textbf{Agentic Internal Intelligence Datasets.}
We summarize the training datasets that aim to enhance agentic internal intelligence capabilities, namely reasoning, reflection, and memory.
MAVIS~\cite{zhang2024mavis} constructs valuable mathematical visual reasoning rationales through automated generation.
R$^3$V~\cite{r3v_reflection} provides 5K response-wise reflection SFT samples annotated by GPT.
LLaVA-CoT~\cite{xu2025llavacotletvisionlanguage} offers 100K structured chain-of-thought SFT samples distilled from GPT-4o.
Mulberry-260K~\cite{yao2024mulberryempoweringmllmo1like} leverages collective MCTS to search 260K reasoning and reflection data.
Vision-R1~\cite{huang2025visionr1} utilizes GPT to generate cold-start data for RL, which contains a substantial amount of reflective content.
R1-Onevision~\cite{yang2025r1onevision} also generates cold-start thinking data from complex visual reasoning tasks.
MMK12~\cite{meng2025mmekura} collects new mathematics problems from textbooks and examination papers ranging from elementary to high school levels.
OpenVLThinker~\cite{deng2025openvlthinkerearlyexplorationcomplex} provides cold-start SFT data and RL data for curriculum-based reinforcement learning.
ThinkLite-VL~\cite{wang2025thinklitevl} repurposes MCTS to identify hard sample for effective RL optimization.
Revisual~\cite{revisual-r1} comprises 47K textual thought samples with reasoning paths, augmented by 31K text and 21K multimodal questions for RL.
GThinker~\cite{zhan2025gthinker} adopts an iterative annotation process to generate 7K reasoning paths for SFT, followed by 4K curated samples for RL.
Video-R1~\cite{feng2025videor1} constructs 165K cold-start SFT samples and 260K RL training samples, both comprising image and video data.
MedTVT-QA~\cite{zhang2025medtvt} is a curated instruction dataset featuring question–answer pairs for physiological interpretation and disease diagnosis using a chain-of-evidence approach.
WeThink~\cite{yang2025wethink} introduces a scalable pipeline that generates context-aware, reasoning-centric QA pairs from images, yielding 120K multimodal QA pairs with annotated reasoning paths.
AVQA-R1-6K~\cite{xing2025Echoink-r1} is a multimodal dataset of synchronized audio–image pairs with multiple-choice questions.
Long-VILA~\cite{chen2024longvila} and Video-XL-pro~\cite{liu2025video-xl-pro} introduce extended long-form video datasets for vision–language fine-tuning and enhance memory modeling.

\textbf{Agentic External Tool Invocation.}
We summarize the training datasets for agentic external tool invocation, covering tasks such as search, code, and visual processing.
Search-R1~\cite{jin2025searchr1}, Search-o1~\cite{li2025search-o1}, and R1-Search~\cite{song2025r1-searcher} contribute text-based reinforcement learning datasets tailored to knowledge-intensive search.
Subsequently, FVQA~\cite{wu2025mmsearchr1} and MAT~\cite{liu2025visualarft} introduce knowledge-intensive multimodal datasets.
These knowledge-intensive, multi-hop datasets are built from challenging and up-to-date knowledge transformed into QA pairs, as exemplified by methods such as WebSailor~\cite{li2025websailor}, WebDancer~\cite{wu2025webdancer}, and AgentFounder~\cite{su2025agenticCPT}.
MathCoder~\cite{wang2023mathcoder} and ReTool~\cite{feng2025retool} provide code datasets for SFT, while ToRL~\cite{li2025torl} and rStar-Coder~\cite{liu2025rstar} construct datasets suitable for reinforcement learning in agentic training. Besides, several projects such as DeepEyes~\cite{zheng2025deepeyes}, Pixel-Reasoner~\cite{su2025pixel}, and Thyme~\cite{zhang2025thyme}, have open-sourced curated datasets for interleaved text-and-image reasoning, which can be used for SFT or RL training.

\textbf{Agentic Environment Interaction.}
We summarize the training datasets for agentic environment interaction, spanning both virtual and physical environments. Specifically, GUI-World~\cite{chen2024gui-world} introduces the first video-based GUI dataset, while Show-UI~\cite{lin2025showui}, GUI-R1~\cite{luo2025guir1}, UI-R1~\cite{lu2025uir1} and GUI-Reflection~\cite{wu2025gui-reflection} provide high-quality image-based alternatives. In embodied AI, the navigation domain is supported by datasets such as VLN-Ego~\cite{qi2025vln} and InternData-N1~\cite{interndata_n1}, whereas the VLA-IT~\cite{yang2025instructvla} dataset serves as a key resource for embodied manipulation.

\subsection{Evaluation Dataset}
\setlength\tabcolsep{24pt}
\label{sec: 5.3 evaluation dataset}

\begin{table*}[ht]
\caption{Summary of datasets for \textbf{evaluating} agentic MLLMs, where T, I, and V represent text, image, and video.}
\centering
\resizebox{0.84\linewidth}{!}{
\begin{tabular}{l l l l l l}
\toprule[1pt]
\rowcolor{red!5}
\textbf{Benchmark} & \textbf{Link}  & \textbf{Type} & \textbf{Scope} & \textbf{Modality} & \textbf{Samples} \\
\midrule
\rowcolor{blue!5}
\multicolumn{6}{c}{\textbf{Agentic Internal Intelligence}} \\
\midrule
MMBench v1.1~\cite{liu2024mmbench} & \href{https://github.com/open-compass/MMBench}{Data} & Reasoning & General & T, I & 3,217 \\

ZeroBench~\cite{roberts2025zerobench} & \href{https://zerobench.github.io/}{Data} & Reasoning & General & T, I & 100 \\

MMMU-Pro~\cite{yue2024mmmu-pro} & \href{https://mmmu-benchmark.github.io/}{Data} & Reasoning & General & T, I & 3,460 \\

MME-CoT~\cite{jiang2025mme_cot} & \href{https://mmecot.github.io/}{Data} & Reasoning & General & T, I & 1,130 \\

M3CoT~\cite{chen2024m3cot} & \href{https://github.com/LightChen233/M3CoT}{Data} & Reasoning & General & T, I & 11,459 \\

ZebraLogic~\cite{lin2025zebralogic} & \href{https://huggingface.co/spaces/allenai/ZebraLogic}{Data} & Reasoning, Reflection & STEM & T, I & 1,000 \\
ZeroBench~\cite{roberts2025zerobench} & \href{https://zerobench.github.io/}{Data} & Reasoning, Reflection & STEM & T, I & 100 \\
OlympiadBench~\cite{he2024olympiadbench} & \href{https://github.com/OpenBMB/OlympiadBench}{Data} & Reasoning, Reflection & STEM & T, I & 8,476  \\
MathVision~\cite{mathvision} & \href{https://huggingface.co/datasets/MathLLMs/MathVision}{Data} & Reasoning, Reflection & STEM & T, I & 3,040 \\
MathVerse~\cite{zhang2024mathverse} & \href{https://github.com/ZrrSkywalker/MathVerse}{Data} & Reasoning, Reflection &  STEM & T, I & 2,612 \\
MMReason~\cite{yao2025mmreason} & \href{https://github.com/HJYao00/MMReason}{Data} & Reasoning, Reflection & STEM & T, I & 2,941\\

WeMath~\cite{qiao2024wemath} & \href{https://github.com/We-Math/We-Math}{Data} & Reasoning, Reflection & STEM & T, I & 6,500\\

VideoMathQA~\cite{rasheed2025videomathqa} & \href{https://mbzuai-oryx.github.io/VideoMathQA/}{Data} & Reasoning, Reflection & STEM & T, V & 2,100 \\

CharXiv~\cite{wang2024charxiv} & \href{https://charxiv.github.io/}{Data} & Reasoning & Chart & T, I & 2,323 \\

LoCoMo~\cite{maharana2024evaluating} &\href{https://github.com/snap-research/locomo}{Data} & Memory  & General &  T, I & 50 \\
MileBench~\cite{song2024milebench} & \href{https://milebench.github.io/}{Data}  & Memory & General & T, I &  6,440 \\

MMLongBench-Doc~\cite{ma2024mmlongbench} & \href{https://mayubo2333.github.io/MMLongBench-Doc/}{Data} & Reasoning, Memory & Doc & T, I & 135 \\
LongVideoBench~\cite{wu2024longvideobench} &  \href{https://github.com/longvideobench/LongVideoBench}{Data} & Reasoning, Memory & General & T, V & 6,678\\
LVBench~\cite{wang2024lvbench} & \href{https://lvbench.github.io/}{Data} & Reasoning, Memory & General & T, V & 1,549 \\

\midrule
\rowcolor{blue!5}
\multicolumn{6}{c}{\textbf{Agentic External Tool Invocation}} \\
\midrule
Humanity's Last Exam~\cite{phan2025humanity_last_exam} & \href{https://agi.safe.ai/}{Data} & Search & General & T, I & 2,500 \\
MM-BrowseComp~\cite{li2025mm-BrowseComp} & \href{https://github.com/MMBrowseComp/MM-BrowseComp}{Data}   & Search & General & T, I & 224 \\
BrowseComp-VL~\cite{geng2025webwatcher} & \href{https://github.com/Alibaba-NLP/DeepResearch/tree/main/WebAgent/WebWatcher}{Data} & Search & General & T, I & 399 \\

FVQA~\cite{wu2025mmsearchr1} & \href{https://huggingface.co/datasets/lmms-lab/FVQA}{Data} & Search & General &  T, I & 1,800   \\
MMSearch~\cite{jiang2024mmsearch} & \href{https://mmsearch.github.io/}{Data} & Search& General & T, I & 300\\
MMSearch-Plus~\cite{tao2025mmsearch} & \href{https://mmsearch-plus.github.io/}{Data} & Search & General & T, I & 311\\
ViDoSeek~\cite{wang2025vidorag} & \href{https://github.com/Alibaba-NLP/ViDoRAG}{Data} & Search & Doc & T, I & 1,200 \\
MAT~\cite{liu2025visualarft} & \href{https://github.com/Liuziyu77/Visual-RFT/tree/main/Visual-ARFT}{Data}& Search, Code & General &  T, I & 350 \\
WebMMU~\cite{awal2025webmmu} & \href{https://webmmu-paper.github.io/}{Data} & Code  & General & T, I & 10,199\\
Design2Code~\cite{si2024design2code} & \href{https://github.com/NoviScl/Design2Code}{Data} & Code & Webpage &  T, I & 484\\
Flame-React-Eval~\cite{Flame-React-Eval} & \href{https://huggingface.co/datasets/Flame-Code-VLM/Flame-Eval-React}{Data} & Code & UI & T, I & 80 \\
V$^*$Bench~\cite{wu2024v} &\href{https://huggingface.co/datasets/craigwu/vstar_bench}{Data} & Visual Processing & General & T, I & 191\\

HRBench~\cite{wang2025divide} &\href{https://github.com/DreamMr/HR-Bench}{Data} & Visual Processing & General & T, I & 200 \\

\midrule
\rowcolor{blue!5}
\multicolumn{6}{c}{\textbf{Agentic Enviroment Interaction}} \\
\midrule

ScreenSpot~\cite{cheng2024seeclick} & \href{https://huggingface.co/datasets/rootsautomation/ScreenSpot}{Data}& Virtual & General & T, I & 1200  \\
ScreenSpot-Pro~\cite{li2025screenspot} & \href{https://gui-agent.github.io/grounding-leaderboard/}{Data} & Virtual & General & T, I & 1,581 \\
AndriodWorld~\cite{rawles2024androidworld} & \href{https://github.com/google-research/android_world}{Data} & Virtual & Andriod & T, I & 116  \\
AndriodControl~\cite{li2024effects} & \href{https://github.com/google-research/google-research/tree/master/android_control}{Data} & Virtual & Andriod &  T, I & 15,283 \\

OSWorld~\cite{xie2024osworld} & \href{https://os-world.github.io/}{Data}  & Virtual & Computer & T, I & 369 \\

WebWalkerQA~\cite{wu2025webwalker} &\href{https://huggingface.co/datasets/callanwu/WebWalkerQA}{Data}  & Virtual & Web & T, I & 680\\ 

OmniACT~\cite{kapoor2024omniact}  & \href{https://huggingface.co/datasets/Writer/omniact}{Data} & Virtual & Web,Desktop & T, I & 9,802 \\

LH-VLN~\cite{lh-vln} & \href{https://hcplab-sysu.github.io/LH-VLN/}{Data} & Physical & Navigation & T, I & 3,260 \\
HA-VLN~\cite{HA-VLN} & \href{https://ha-vln-project.vercel.app/}{Data} & Physical & Navigation & T, I & 16,844 \\
VLABench~\cite{zhang2024vlabench} & \href{https://vlabench.github.io/}{Data} & Physical & Manipulation & T, I & 2,164 \\

\bottomrule[1pt]
\end{tabular}
}
\label{tab:agentic-MLLM-bench}
\end{table*}

We survey the benchmarks used to evaluate the agentic capabilities of MLLMs, as presented in Table~\ref{tab:agentic-MLLM-bench}.

\subsubsection{Benchmark Internal Intelligence} 
\begin{itemize}
    \item \textbf{Benchmark Reasoning and Reflection Capabilities.}
    (1) General Problems. We review recent benchmarks for general visual question answering that are relatively more challenging and require reasoning, including MMBench v1.1~\cite{liu2024mmbench}, M3CoT~\cite{chen2024m3cot}, MME-CoT~\cite{jiang2025mme_cot}, and MMMU-Pro~\cite{yue2024mmmu-pro}.
    (2) STEM Problems. Science, technology, engineering, and mathematics (STEM) problems are more challenging and complex, requiring MLLMs to possess stronger long-chain reasoning and reflective capabilities in order to solve them effectively, including MathVision~\cite{mathvision}, MathVerse~\cite{zhang2024mathverse}, OlympiadBench~\cite{he2024olympiadbench}, MMReason~\cite{yao2025mmreason}, WeMath~\cite{qiao2024wemath}, and VideoMathQA~\cite{rasheed2025videomathqa}.
    (3) Chart and Document Problems. Chart and document problems require cross-modal alignment and numerical reasoning, as illustrated by benchmarks such as CharXiv~\cite{wang2024charxiv} and MMLongBench-Doc~\cite{ma2024mmlongbench}.
    \item \textbf{Benchmark Memory Capabilities.}
    Evaluating the memory capabilities of MLLMs focuses on their ability to retain and utilize information over long multi-modal contexts and multi-turn conversations. Benchmarks in this category include MileBench~\cite{song2024milebench}, MMLongBench-Doc~\cite{ma2024mmlongbench}, LongVideoBench~\cite{wu2024longvideobench}, and LVBench~\cite{wang2024lvbench}, which assess how well models can preserve contextual information, recall relevant details, and maintain coherent reasoning across extended interactions.
\end{itemize}

\subsubsection{Benchmark External Tool Invocation.} 
\begin{itemize}
    \item \textbf{Benchmark Search Capabilities.}
    Evaluating agentic search capabilities typically relies on benchmarks composed of multi-hop, knowledge-intensive, and up-to-date questions. Such tasks require the model not only to retrieve relevant information from external resources but also to integrate evidence across multiple sources and reason over them to reach a correct conclusion. Representative benchmarks include Humanity’s Last Exam~\cite{phan2025humanity_last_exam}, MM-BrowseComp~\cite{li2025mm-BrowseComp}, BrowseComp-VL~\cite{geng2025webwatcher}, FVQA~\cite{wu2025mmsearchr1}, MMSearch~\cite{jiang2024mmsearch}, MMSearch-Plus~\cite{tao2025mmsearch}, ViDoSeek~\cite{wang2025vidorag}, and MAT~\cite{liu2025visualarft}.
    \item \textbf{Benchmark Code Capabilities.} 
    Code benchmarks evaluate how well MLLMs can generate code across multiple languages, \eg Python, JavaScript, and SQL.
    Representative benchmarks include WebMMU~\cite{awal2025webmmu}, Design2Code~\cite{si2024design2code} and Flame-React-Eval~\cite{Flame-React-Eval}. Additionally, several advanced mathematical benchmarks, such as AIME2024 and AIME2025, are commonly employed to evaluate the code-integrated reasoning capabilities.

    \item \textbf{Benchmark Visual Processing Capabilities.} Benchmarks for high-resolution image understanding (e.g., V*Bench~\cite{wu2024v} and HRBench~\cite{wang2025divide}) evaluate the visual processing capability that requires agentic MLLMs to perform operations like cropping and zooming to uncover visual clues, leading to enhanced image comprehension.
\end{itemize}

\subsubsection{Benchmark Environment Interaction}

\begin{itemize}
    \item \textbf{Benchmark Virtual Interaction Capabilities.} A range of GUI benchmarks, such as ScreenSpot~\cite{cheng2024seeclick}, AndroidWorld~\cite{rawles2024androidworld} and OSWorld~\cite{xie2024osworld}, serve to evaluate virtual interaction capabilities. These benchmarks provide diverse environments where agents must execute tasks by interacting with graphical user interfaces, testing their ability to understand screen elements and perform correct sequences of actions. 

     \item \textbf{Benchmark Physical Interaction Capabilities.} In embodied AI and robotics, core physical interaction capabilities are evaluated across key domains, with navigation assessed on benchmarks such as LH-VLN~\cite{lh-vln} and HA-VLN~\cite{HA-VLN}, and manipulation evaluated using VLABench~\cite{zhang2024vlabench}.
\end{itemize}

\section{Application}
\label{sec: 6 application}
Agentic MLLMs, endowed with strong generalization capabilities and integrated agentic functionalities, have demonstrated remarkable potential across a broad spectrum of downstream tasks.
Unlike previous MLLM agents that are often restricted to specific domains, agentic MLLMs can reason, reflect, leverage memory, invoke various external tools, and interact with dynamic environments, enabling them to handle complex real-world scenarios. 
This transformative paradigm has garnered growing attention from diverse research communities, offering fresh insights into long-standing challenges and unlocking new opportunities for practical applications in areas such as Deep Research, Embodied AI, Healthcare, GUI Agents, Autonomous Driving, and Recommender Systems. 
In the following subsections, we present an overview of these representative applications and highlight how agentic MLLMs are reshaping them.

\subsection{Deep Research}
\label{sec: 6.1 deep research}

Deep Research (DR) represents a milestone in agentic intelligence, showcasing the ability of MLLMs to autonomously conduct multi-step, goal-directed research for high-intensity knowledge work.
Unlike conventional models that rely on single-turn retrieval or user-driven prompting, Deep Research integrates multi-step reasoning and tool use to automate information discovery and synthesis, thereby assisting domains such as finance, science, policy, and education in handling complex tasks~\cite{openai2025deepresearch,xu2025comprehensive,huang2025deep,zhang2025web,zheng2025deepresearcherscalingdeepresearch}. 
Recently, a variety of Deep Research agents have emerged, including OpenAI Deep Research~\cite{openai2025deepresearch}, Gemini Deep Research~\cite{google2025deepresearch}, Grok DeepSearch~\cite{xai2025deepsearch}, Perplexity Deep Research~\cite{perplexity2025deepresearch}, Copilot Researcher~\cite{microsoft2025researcher}, Kimi-Researcher~\cite{moonshot2025kimiresearcher}, AutoGLM~\cite{zhipu2025autoglm}, Tongyi Deep Research~\cite{tongyidr}, MiroThinker~\cite{2025mirothinker}, and Manus~\cite{manus2025}.
These Deep Research systems demonstrate strong capabilities in open-ended, knowledge-intensive tasks, enabling them to tackle the kinds of complex, real-world problems that people encounter in both professional and everyday contexts. 
It thus marks a significant step toward practical, autonomous AI systems capable of scalable and verifiable research.

\subsection{Embodied AI}
\label{sec: 6.2 Embodied AI}
Embodied AI marks a transformative shift from passive perception to active engagement in physical environments, with vision-language-action (VLA) models emerging as a pivotal architectural framework~\cite{kim2024openvla,qu2025embodiedonevision,bu2025univla,lv2025f1}. These models integrate multimodal reasoning with motion control to translate high-level linguistic and visual inputs into executable action sequences, thereby serving as the cognitive core for next-generation robotic systems. In robotics, VLA-powered agents demonstrate remarkable open-world generalization~\cite{black2024pi_0,cheang2025gr,black2025pi0,team2025gemini,lee2025molmoact}, significantly expanding the scope of complex tasks achievable by machines. Beyond technical advancement, this synergy drives substantial commercial value across logistics, smart manufacturing, and personalized service domains, offering scalable, intelligent solutions for dynamic real-world applications~\cite{perlo2025embodied,liu2025aligning}.

\subsection{Healthcare}
\label{sec: 6.3 healthcare}
The rapid advancement of MLLMs has spurred growing interest in their application within healthcare contexts. Unlike general domains, medicine requires exceptional reliability, strict control of hallucinations, and robust interpretability. Early approaches such as LLaVA-Med~\cite{li2023llava} and HuatuoGPT series~\cite{zhang2023huatuogpt,chen2023huatuogpt-ii,chen2024huatuogpt-v} rely on SFT with curated medical QA data, but often exhibit limited generalization. Subsequent efforts like HuatuoGPT-o1 incorporate RL (e.g., PPO) to activate reasoning and self-reflection, markedly improving diagnostic accuracy~\cite{chen2024huatuogpt-o,hao2025surgery,zhang2025medtvt}. Beyond enhancing intrinsic model capabilities, systems such as MMed-RAG~\cite{xia2024mmed} and MedResearcher-R1~\cite{yu2025medreseacher} further integrate external tools like domain-aware retrieval and medical knowledge graphs. These agentic MLLMs combine sophisticated retrieval mechanisms or other advanced tools with RL to achieve state-of-the-art performance on complex medical reasoning tasks~\cite{zhang2025patho,li2025cxr,lan2025gem}. Moreover, medical embodied AI systems~\cite{liu2025screens}, such as those in surgical robotics~\cite{su2024fully,pore2021safe,liu2023robotic}, are also show promising application prospects and practical value, further extending the impact of agentic MLLMs into physical clinical interventions.

\subsection{GUI Agents}
\label{sec: 6.4 GUI_agents}
GUI agents represent a breakthrough application of agentic MLLMs, fundamentally reshaping human-computer interaction~\cite{lu2025uir1,luo2025guir1,qin2025ui,wang2025ui,ye2025mobile}.
They demonstrate remarkable capability in automating complex digital tasks across diverse software environments and operating systems, including web scenarios~\cite{wei2025webagent}, mobile platforms~\cite{ye2025mobile}, and desktop interfaces~\cite{nayak2025ui,xie2024osworld}. By visually perceiving the screen, comprehending natural language commands, and executing precise low-level actions (e.g., clicks, typing, scrolling), they enable a wide range of sophisticated applications, including fundamental tasks like file management and web operations~\cite{zhao2025worldgui}, and more advanced capabilities from cross-app workflow orchestration~\cite{lu2024gui} to personalized user support~\cite{yang2025fingertip}. The advancement of GUI agents holds significant potential to enhance digital accessibility and operational efficiency, thereby offering substantial benefits to both commercial ecosystems and broader societal infrastructures.

\subsection{Autonomous Driving}
\label{sec: 6.5 autonomous_driving}
The application of agentic MLLMs in autonomous driving represents a rapidly evolving research frontier aimed at enhancing complex decision-making and interaction capabilities~\cite{li2025drive,jiang2025alphadrive,cui2025chain,qian2025agentthink}. One line of work incorporates CoT reasoning into autonomous driving systems, utilizing the sophisticated cognitive capabilities of MLLMs to generate accurate and interpretable motion trajectories~\cite{li2025drive,jiang2025alphadrive,yuan2025autodrive,cui2025chain,ishaq2025drivelmm}. Another category of methods integrates external tools, such as object detection, depth estimation, and occupancy prediction, to enhance perceptual robustness and situational awareness. Through SFT combined with RL training, such models learn to autonomously invoke and leverage these tools, significantly improving the robustness and generalization of driving policies in open-world scenarios~\cite{qian2025agentthink}. Together, these efforts highlight a clear trend toward building more reliable, transparent, and tool-augmented MLLM-based agents for autonomous driving. By combining internal reasoning capability with external perceptual tools and advanced training paradigms, agentic MLLMs are poised to overcome key challenges in real-time decision-making, safety assurance, and scalable deployment in dynamic driving environments.

\subsection{Recommender System}
\label{sec: 6.6 Recommender System}
Traditional MLLM-based recommender systems~\cite{ye2025harnessing, liu2024rec} primarily enhance existing recommendation pipelines by leveraging multimodal representations and language understanding to improve ranking, retrieval, and conversational interactions.
However, these systems typically remain reactive: they rely on pre-defined objectives (e.g., click-through rate prediction), static user profiles, and limited dialogue rounds to refine outputs. While MLLMs enable richer modeling of user intent and item semantics, they still lack deeper autonomy and adaptability.
Recently, agentic MLLM recommender systems (MLLM-ARS)~\cite{LLM-ARS, liu2025recoworld, chen2025vragent-r1, zhang2025reasonrec, huang2025towards_agentic_recommender_systems} have emerged to transcend this paradigm by embedding reasoning, reflection, memory, tool use, and virtual interaction within the recommendation process.
Rather than passively responding to user requests, agentic recommenders proactively explore user preferences, simulate future behaviors, and adapt strategies over time. They integrate multimodal cues with agentic capabilities such as reasoning, reflection, and role-playing to deliver interactive, context-aware, and personalized experiences.
Crucially, these systems evolve dynamically, balancing immediate feedback with long-horizon personalization, paving the way for recommender systems that are not only responsive but also autonomous, transparent, and continuously self-improving.

\section{Challenges and Future Directions}
\label{sec: 7 challenges and future directions}
Despite recent progress, the development of agentic MLLMs is still in its early stages, and many challenges remain to be addressed.
This section discusses these limitations and outlines potential directions for future research.

\subsection{Richer Action Space of Agentic MLLM}
Agentic MLLMs have demonstrated remarkable capabilities in handling complex tasks.
However, the action space of existing models, and the range of tools they can access is often restricted to a single type~\cite{wu2025mmsearchr1, feng2025retool}.
Recent studies have integrated a wider range of tool usage. For example, Visual-ARFT~\cite{liu2025visualarft} can perform both search and code execution, while WebWatcher~\cite{geng2025webwatcher} supports even richer functionalities, including search, code interpretation, and internal OCR.
Looking ahead, future agentic MLLMs are expected to operate with a richer action space, equipped to invoke a broader spectrum of external tools and services. They may seamlessly integrate with data analysis platforms, simulation environments, multimodal sensors, and interactive APIs, enabling more adaptive and generalizable agentic behaviors across diverse real-world scenarios.

\subsection{Efficient Agentic MLLMs}
While agentic MLLMs excel at handling complex problems through multi-turn reasoning and external tool invocation, these iterative processes substantially increase their computational and reasoning overhead. 
In some cases, models may require up to thirty minutes to complete a single task~\cite{openai2025deepresearch}, imposing significant costs on both training and inference. 
Such inefficiency poses challenges for real-time applications and large-scale deployment, where latency, energy consumption, and resource constraints become critical considerations. 
Although some studies have accelerated long-chain reasoning~\cite{luo2025o1-pruner, luo2025ada-r1, xiao2025fast, lou2025adacot}, research on speeding up tool invocation remains limited.
To address these issues, future research should focus on improving the efficiency of agentic MLLMs, accelerating both training and inference without compromising performance. 
By enhancing computational efficiency, agentic MLLMs can move closer to practical, scalable deployment across diverse real-world environments.

\subsection{Long-term Agentic Memory}
Long-term memory allows agentic MLLMs to plan, reason, and interact in ways that support continuity, adaptation, and long-term experience accumulation over time.
Although recent studies have explored agentic memory~\cite{ding2024longrope, xu2025a-mem, yan2025memoryr1, zhou2025mem1}, most of these works have focused primarily on the language modality, with limited exploration of multimodal settings.
At the same time, the effective length of memory in current systems remains highly constrained, restricting their ability to sustain coherent knowledge across longer time horizons.
Future work should design persistent memory architectures that allow models to accumulate, organize, and retrieve knowledge across extended time spans.
Such memory must be both scalable, capable of processing the vast multimodal streams agents encounter, and selective, able to filter, compress, and prioritize experiences relevant for reasoning, ultimately supporting evolving memory systems that foster personalization, sustained collaboration, and adaptive problem-solving.
Ultimately, long-term agentic memory is not just a technical refinement but a prerequisite for creating enduring partners capable of continuous learning and alignment with human goals.

\subsection{Agentic Training and Evaluation Dataset}
Currently, the development of agentic MLLMs is still at a very early and exploratory stage, and one of the most pressing challenges lies in the scarcity of training datasets specifically designed for agentic behaviors.
Tongyi Lab~\cite{su2025agenticCPT, wu2025webwalker, tao2025webshaper} proposes a fully automated pipeline for generating synthetic agentic trajectories, supporting CPT, SFT, and RL.
However, much of this data remains inaccessible to the research community and lacks sufficient exploration in multimodal domains. 
Therefore, an urgent research direction lies in developing effective and efficient methods for synthesizing high-quality multimodal agentic trajectory data.
In addition, to evaluate the performance of agentic MLLMs, several recent benchmarks have been established, such as MM-BrowseComp~\cite{li2025mm-BrowseComp} and BrowseComp-VL~\cite{geng2025webwatcher}. However, these benchmarks primarily focus on specific aspects of agentic behavior, while certain actions, such as memory utilization and the ability to coordinate reasoning across multiple tool invocations, still lack effective evaluation datasets. Moreover, robust methods for assessing whether actions are correctly executed remain underexplored.

\subsection{Safe Agentic MLLMs}
AI safety has long been recognized as a central challenge, and prior work~\cite{gu2024mllmguard,ying2024safebench,liu2024mm-safetybench,yang2025mla-trust, pi2024mllm-protector} has focused extensively on building systems that are safe and controllable.
As agentic MLLMs become increasingly autonomous in planning, tool invocation, and environment interaction, ensuring their safety will be a critical research priority.
Unlike static models, agentic systems dynamically generate action sequences that may call external tools, APIs, or even physical devices, thereby amplifying the risks of unintended consequences~\cite{raza2025trism}.
For instance, a model conducting web search may retrieve incorrect or harmful information, which can bias subsequent MDP-based decision making and lead to unsafe downstream actions.
In multimodal settings, the difficulty is further magnified, as ambiguous or adversarial inputs can propagate across modalities and destabilize agent behavior. Addressing these challenges requires a combination of rigorous benchmarking, adversarial stress-testing, and the integration of normative frameworks, ultimately ensuring that agentic MLLMs remain reliable, controllable, and aligned with human intent as they advance toward more general autonomy.

\section{Conclusion}
This survey charts the recent advances of agentic MLLMs, marking a pivotal shift from traditional MLLM agents to models with agentic capabilities.
We begin by discussing MLLM agents and agentic MLLMs, the latter distinguished by dynamic workflows, proactive execution of actions, and strong generalization across domains.
We then introduce agentic foundational MLLMs, action space, CPT, SFT, RL, and evaluation methodologies, which together serve as the preliminary knowledge base.
Building on it, we propose a threefold taxonomy that organizes MLLM agentic capabilities into: (i) internal intelligence, where reasoning, reflection, and memory coordinate long-horizon decisions; (ii) external tool invocation, where models proactively call search engines, code executors, and visual processing to acquire and manipulate information; and (iii) environment interaction, where agents act within virtual and physical settings to obtain feedback and continuously refine their plans through iterative adaptation.
In addition, we consolidated open-source training frameworks, training datasets, and evaluation benchmarks to provide a practical reference that can ground and accelerate future research, and we summarized emerging agentic applications across diverse scenarios.
We also track notable developments through a real-time \href{https://github.com/HJYao00/Awesome-Agentic-MLLMs}{GitHub repository} and hope that these resources will help accelerate the advancement of agentic MLLMs.


%





\ifCLASSOPTIONcaptionsoff
  \newpage
\fi



%




{\small
\bibliographystyle{revision_ref}
\bibliography{ref}
}

%








\end{document}